\documentclass{article}

\usepackage[preprint]{neurips_2024}

\usepackage[utf8]{inputenc} 
\usepackage[T1]{fontenc}    
\usepackage{hyperref}       
\usepackage{url}            
\usepackage{booktabs}       
\usepackage{amsfonts}       
\usepackage{nicefrac}       
\usepackage{microtype}      
\usepackage{xcolor}         
\usepackage{amsmath}
\usepackage{amsfonts}
\usepackage{multirow}
\usepackage{booktabs}
\usepackage{hyperref}
\usepackage{graphicx}
\usepackage{times}
\usepackage{latexsym}
\usepackage{colortbl}
\usepackage{xcolor}
\usepackage[utf8]{inputenc}
\usepackage{amssymb}
\usepackage{multirow}
\usepackage{bigdelim}
\usepackage{todonotes}
\usepackage{longtable}
\usepackage{tabularray}
\usepackage{wrapfig}
\usepackage[most]{tcolorbox} 
\usepackage{url}
\usepackage{xspace}
\usepackage{svg}
\usepackage[absolute]{textpos} %

\usepackage{fdsymbol}   %

\newcommand{\huggingface}{\raisebox{-1.5pt}{\includegraphics[height=1.05em]{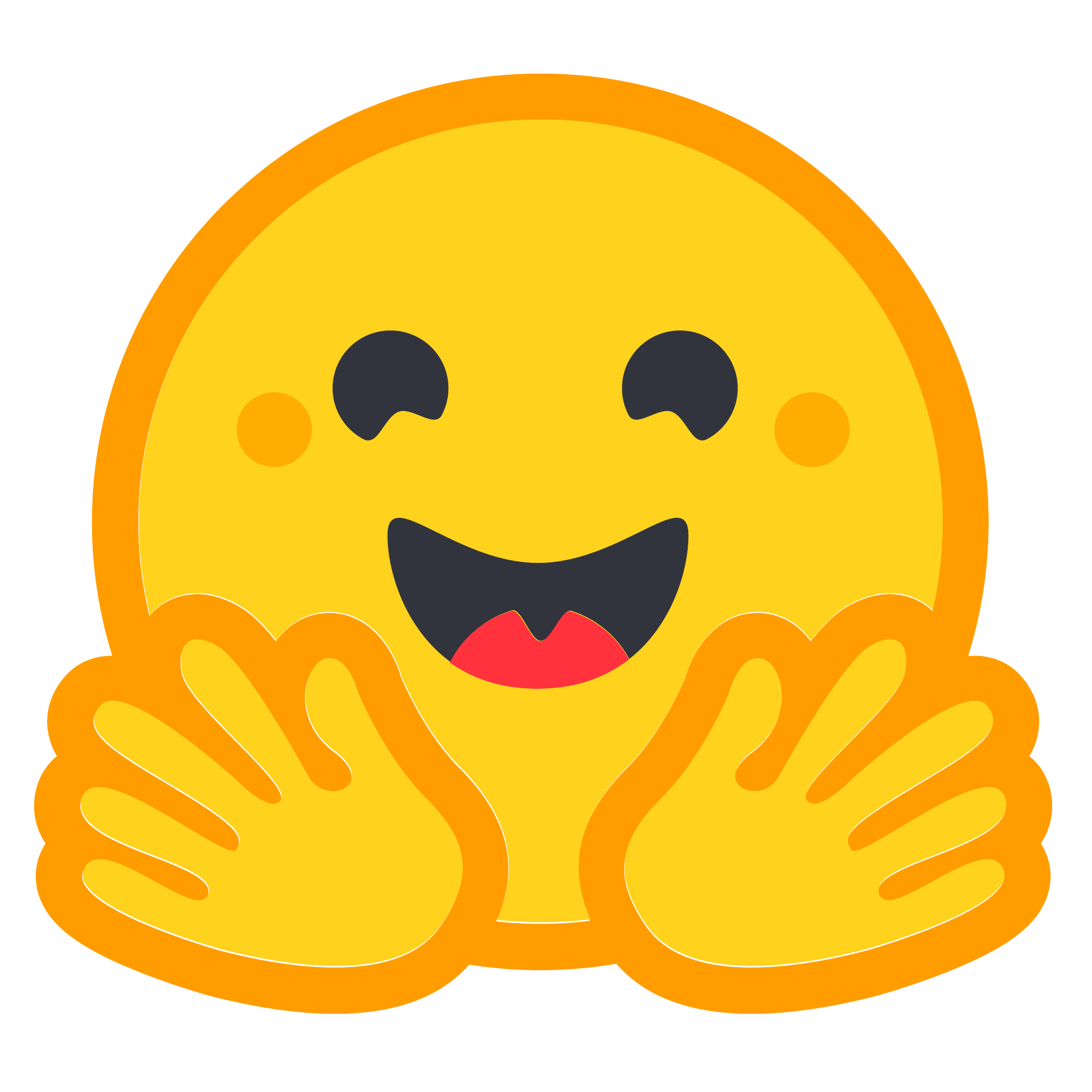}}\xspace}

\newcommand{\hfdataset}{\raisebox{-1.5pt}{\includegraphics[height=1.05em]{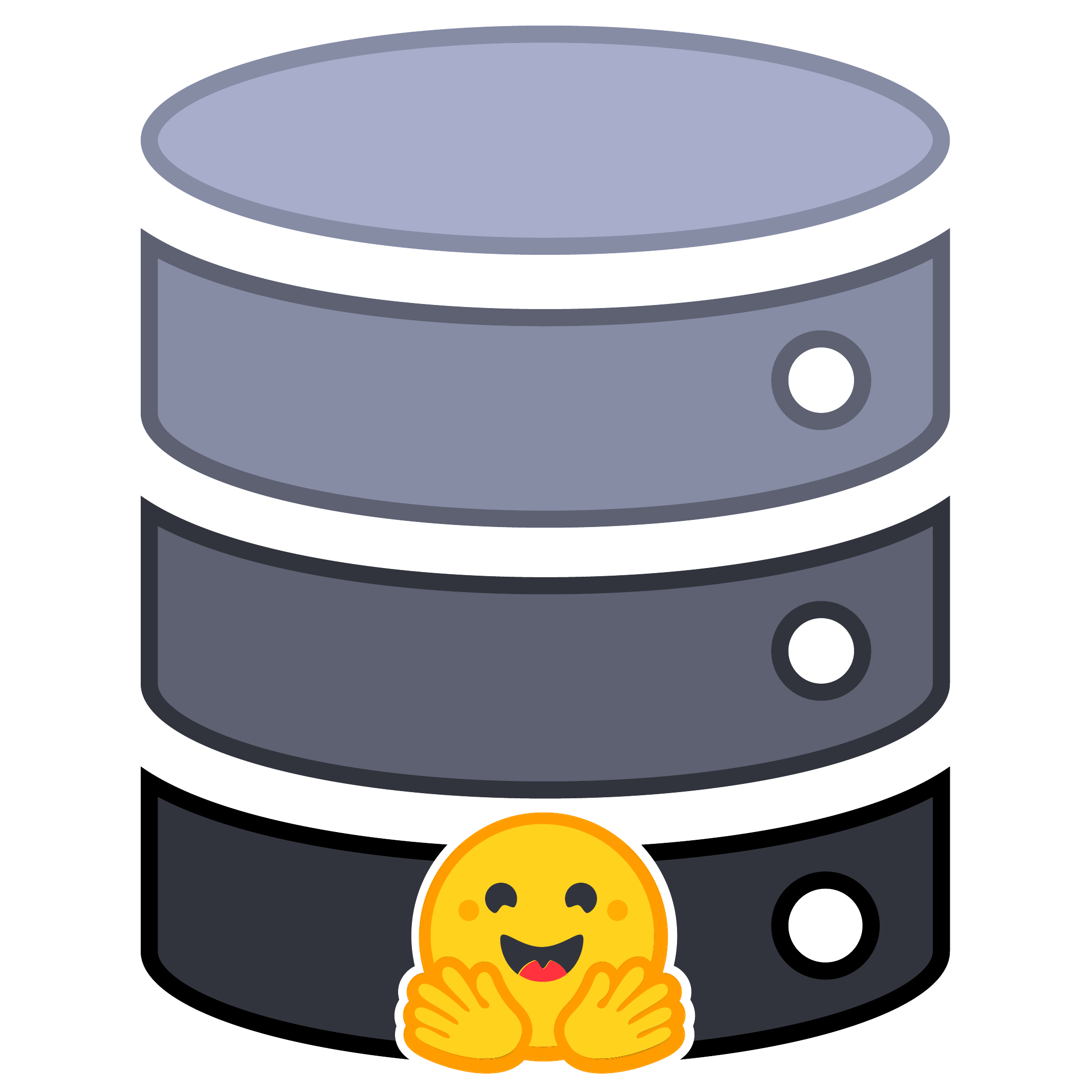}}\xspace}
\newcommand{\github}{\raisebox{-1.5pt}{\includegraphics[height=1.05em]{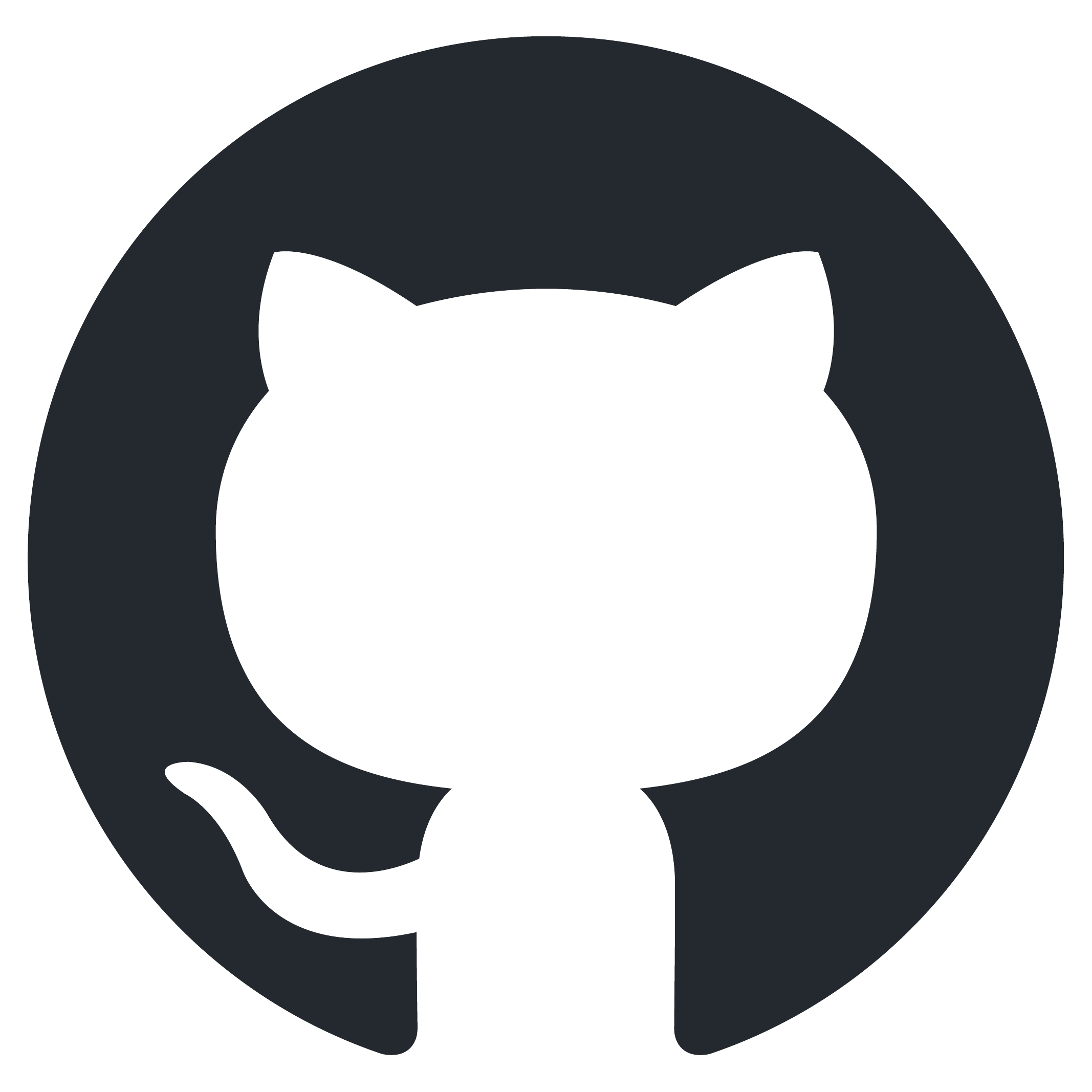}}\xspace}

\newcommand{\ar}{\raisebox{-1.5pt}{\includegraphics[height=1.05em]{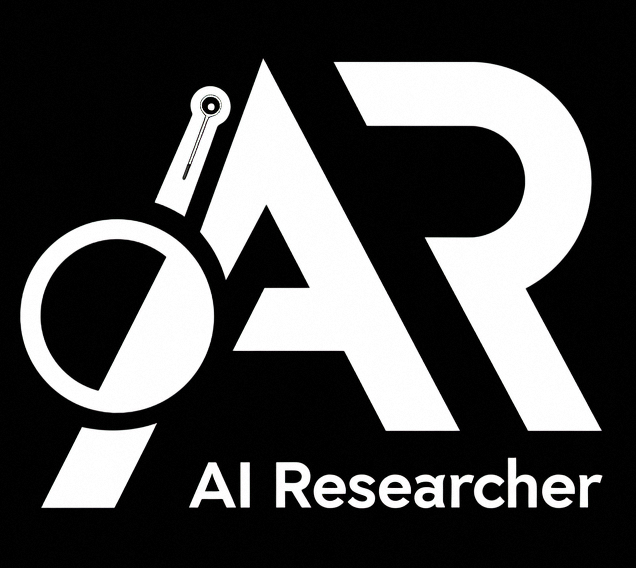}}\xspace}

\usepackage{microtype}
\usepackage{fontawesome5}  
\usepackage{inconsolata}
\usepackage{academicons}     

\usepackage{graphicx}
\title{DeepReview: Improving LLM-based Paper Review with Human-like Deep Thinking Process
}

\author{
  \textbf{Minjun Zhu} \\
  Zhejiang University\\
  School of Engineering, Westlake University\\
  \texttt{zhuminjun@westlake.edu.cn} \\
  \And
  \textbf{Yixuan Weng} \\
  School of Engineering, Westlake University\\
  Research Center for Industries of the Future\\
  \texttt{wengsyx@gmail.com} \\
  \And
  \textbf{Linyi Yang} \\
  University College London\\
  \texttt{yanglinyiucd@gmail.com} \\
  \And
  \textbf{Yue Zhang}\thanks{Corresponding Author. Supported by Research Center for Industries of the Future, Westlake University.} \\
  School of Engineering, Westlake University\\
  Research Center for Industries of the Future\\
  \texttt{zhangyue@westlake.edu.cn}
}

\begin{document}

\maketitle

\begin{abstract}

 Large Language Models (LLMs) are increasingly utilized in scientific research assessment, particularly in automated paper review. However, existing LLM-based review systems face significant challenges, including limited domain expertise, hallucinated reasoning, and a lack of structured evaluation. To address these limitations, we introduce DeepReview, a multi-stage framework designed to emulate expert reviewers by incorporating structured analysis, literature retrieval, and evidence-based argumentation. Using DeepReview-13K, a curated dataset with structured annotations, we train DeepReviewer-14B, which outperforms CycleReviewer-70B with fewer tokens. In its best mode, DeepReviewer-14B achieves win rates of 88.21\% and 80.20\% against GPT-o1 and DeepSeek-R1 in evaluations. Our work sets a new benchmark for LLM-based paper review, with all resources publicly available. The code, model, dataset and demo have be released in \url{http://ai-researcher.net}.
\end{abstract}
\section*{Resources}

\quad\huggingface Models: \quad{\href{https://huggingface.co/WestlakeNLP/DeepReviewer-7B}{\texttt{DeepReviewer-7B}} \quad \href{https://huggingface.co/WestlakeNLP/DeepReviewer-14B}{\texttt{DeepReviewer-14B}}}

\vspace{.5em}\quad\hfdataset Dataset: \quad{\href{https://huggingface.co/WestlakeNLP/DeepReview-13K}{\texttt{DeepReview-13K}} }

\vspace{.5em}\quad\github Code Repository: \quad{\href{https://github.com/zhu-minjun/Researcher}{\texttt{zhu-minjun/Researcher}}  }

\vspace{.5em}\quad\ar HomePage: \quad{\href{http://ai-researcher.net}{\texttt{ai-researcher.net}}}

\vspace{.5em}\quad\ar Demo: \quad{\href{http://ai-researcher.net/deepreviewer}{\texttt{ai-researcher.net/deepreviewer}}}

\vspace{.5cm}
\textcolor{red}{ \normalsize Claim: This work is not advocating LLM replacement of human reviewers but rather exploring LLM assistance in peer review.}
\section{Introduction}
Peer review is the foundation of scientific progress, ensuring that research is novel, reliable, and rigorously evaluated by experts before publication \citep{alberts2008reviewing}. With the increasing volume of research submissions, Large Language Models (LLMs) have become promising tools to support reviewers~\citep{yang-etal-2024-large-language,chris2024the,li2024automated,scherbakov2024emergencelargelanguagemodels,si2025can}. For example, the ICLR 2025 conference has introduced an LLM-based system to assist reviewers in providing feedback \cite{iclr2025}. 

Recent research has explored two primary approaches to improve LLM-based review systems: (1) employing LLM-powered agents to simulate the peer review process, as exemplified by AI-Scientist \citep{chris2024the} and AgentReview \citep{jin-etal-2024-agentreview}; and (2) developing open-source models trained on extensive datasets from existing peer review platforms, such as ReviewMT \citep{tan2024peerreviewmultiturnlongcontext} and CycleReviewer \citep{yixuan2024cycleresearcher}.


Despite these advancements, current systems exhibit several critical limitations: they struggle to comprehensively identify submission flaws, resulting in superficial feedback \citep{zhou-etal-2024-llm}; lack evidence-based justifications \citep{zhuang2025large}; and fail to provide clear, actionable suggestions \citep{ye2024we,du2024llms}. Moreover, their vulnerability to prompt engineering leads to inaccurate evaluations \citep{ye2024we}. While robust feedback is crucial for scientific advancement and peer review integrity, developing reliable evaluation frameworks faces two significant challenges: (1) The scarcity of structured paper review datasets that capture fine-grained expert evaluation processes. Most available open review datasets primarily contain aggregated reviews and decisions, limiting LLMs' ability to learn systematic review reasoning chains and increasing their susceptibility to shortcut learning and adversarial manipulation. (2) LLMs' inherent constraints, including restricted domain knowledge, lack of dynamic knowledge updating mechanisms, and a tendency to generate hallucinated content without adequate verification \citep{schintler2023criticalexaminationethicsaimediated,drori2024human}, which significantly impair their capability to assess complex scientific content \citep{wang-etal-2020-reviewrobot,yuan2021automatescientificreviewing}. 


To address these challenges, we introduce \textbf{DeepReview}, a structured multi-stage review framework that closely aligns with the expert review process by incorporating novelty assessment, multi-dimensional evaluation criteria, and reliability verification. We develop a comprehensive data synthesis pipeline that integrates retrieval and ranking\citep{asai2024openscholarsynthesizingscientificliterature}, self-verification \citep{weng2023large}, and self-reflection \citep{ji-etal-2023-towards}, ensuring the soundness and robustness of LLM-generated suggestions. This approach enables deeper insights into the reasoning and decision-making of paper review. The resulting dataset, \textbf{DeepReview-13K}, consists of raw research papers, structured intermediate review steps, and final assessments. Based on that, we train \textbf{DeepReviewer-14B}, a model that offers three inference modes -- Fast, Standard, and Best -- allowing users to balance efficiency and response quality. We further construct \textbf{DeepReview-Bench}, a comprehensive benchmark containing 1.2K samples, which evaluates both quantitative aspects (rating prediction, quality ranking, and paper selection) and qualitative review generation through LLM-based assessment.

Extensive experiments demonstrate DeepReviewer 14B's superior performance across multiple dimensions. Compared to existing systems like  CycleReviewer-70B, GPT-o1, and Deepseek-R1, our model achieves substantial improvements in Score (Rating MSE: 44.80$\%\uparrow$), Ranking (Rating Spearman: 6.04$\%\uparrow$), and Selection (Accuracy 1.80$\%\uparrow$). In LLM-as-a-judge evaluation \citep{wang2024pandalm,rewina2025potential}, it achieves a 80\% win rate against GPT-o1 and Deepseek-R1. Notably, DeepReviewer exhibits strong resilience to adversarial attacks despite no explicit robustness training. Furthermore, our Test-Time Scaling analysis reveals that DeepReviewer can enhance its performance by adjusting reasoning paths and response lengths. 

Our work establishes a foundation for robust LLM-based review systems through DeepReview, a structured framework that addresses fundamental challenges in automated manuscript evaluation. We introduce DeepReview-13K, a dataset featuring fine-grained review reasoning chains, alongside DeepReview-Bench, a benchmark for automated paper review. Built upon these resources, our DeepReviewer-14B model demonstrates substantial improvements over existing approaches while maintaining strong resilience to adversarial attacks, validating the effectiveness of our structured approach to automated scientific evaluation. Our code, model, and data will be publicly available under the agreement of our usage policy.


\section{Related Work}

\textbf{Reasoning in LLMs.} The emergence of large language models \cite{achiam2023gpt,touvron2023llama,bai2023qwen} has provided new assistance in advancing solutions to complex Science challenges \cite{hendrycksmath2021}. Initially, Scratchpads and chain-of-thought \cite{akyurek2022learning,nye2021show,weichain} encouraged LLMs to think. This technique has been employed in various reasoning tasks. Building on this, a series of works including self-consistency \cite{wangself}, self-verification \cite{weng2023large}, and self-reflection \cite{madaan2024self} prompted language models to output more thinking processes during reasoning. Later, OpenAI's O1 model \cite{jaech2024openai} and various open-source long chain-of-thought models \cite{guo2025deepseek} achieved Scaling Test-time Compute \cite{yao2024tree,guan2025rstarmathsmallllmsmaster} through additional supervised training or reinforcement learning, enabling language models to select optimal solutions for improved performance \cite{xiang20252reasoningllmslearning}. While these advances have enhanced reasoning capabilities, they primarily focus on general problem-solving rather than specialized academic review tasks. Our Review-with-Thinking framework extends these reasoning approaches specifically for peer review.

\textbf{Reliable Scientific Literature Assessment.} Recent studies have demonstrated significant progress in automated scientific research. \citep{chris2024the} develop an AI scientist for autonomous hypothesis generation and experimentation \citep{langley1987scientific,daniil2023emergent,Aider,zonglin2023large,li-etal-2024-simulating,hu2024nova}. Multi-agent frameworks \citep{ghafarollahi2024sciagents, rasal2024navigating, su2024two} enable collaborative scientific reasoning, while \citep{yixuan2024cycleresearcher} show LLM-based review systems can enhance scientific discovery through reinforcement learning. However, these systems often lack structured reasoning, resulting in unreliable feedback.

\textbf{Robust LLM-based Paper Review.}
Recent work spans generation-focused approaches using role-playing agents \citep{d2024marg, gao2024reviewer2, yu2024automated,yixuan2024cycleresearcher}, meta-review synthesis \citep{santu2024prompting, li2023summarizing, zeng2024scientific}, and bias detection mechanisms \citep{liang2024monitoring, tyser2024ai, tan2024peer}. Hybrid workflows \citep{jin2024agentreview, zyska2023care} combine human-AI collaboration with iterative refinement. While evaluation benchmarks \citep{funkquist2022citebench, zhou2024llm, kang2018dataset} and ethical analyses \citep{ye2024we, latona2024ai} have advanced the field, existing systems struggle with complex assessments and remain vulnerable to adversarial attacks, highlighting the need for explicit reasoning processes.

\begin{figure*}[t]
    \centering
    \includegraphics[width=\textwidth]{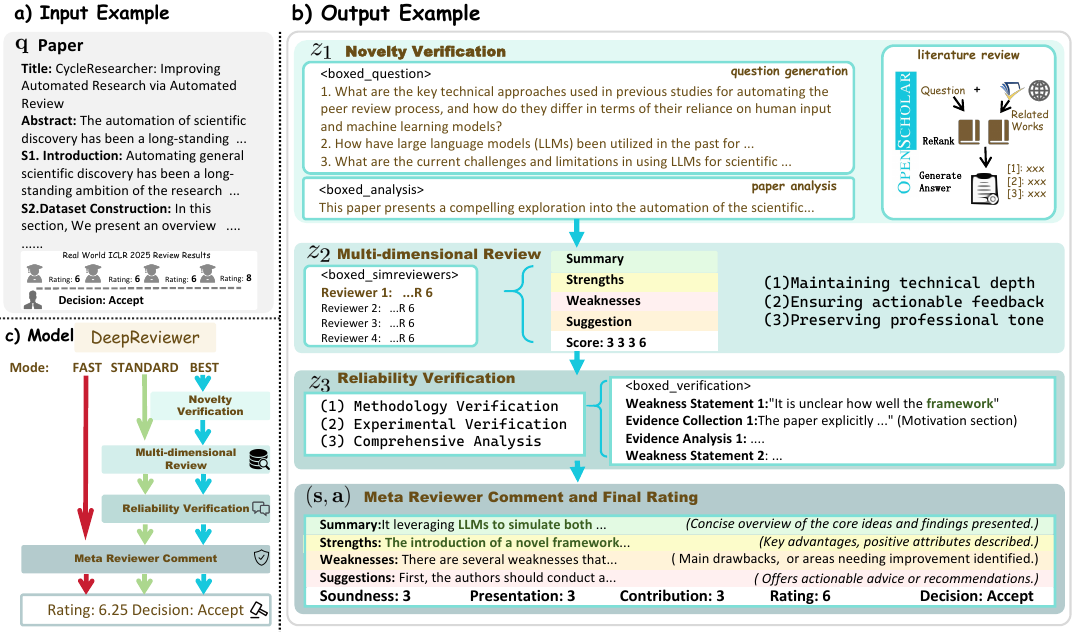}
    \vspace{-0.7cm}
    \caption{Overview of the DeepReviewer. (a) Input paper example with a real-world research paper. (b) Output example showing DeepReviewer's multi-stage reasoning process: Novelty Verification, Multi-dimension Review, and Reliability Verification. (c) Inference modes: fast, standard, and best, highlighting different reasoning paths. We provide a more detailed case study in the appendix \ref{appendix:case}.}
    \vspace{-0.3cm}
    \label{fig:main}
\end{figure*}

\section{Data Collection}



We present DeepReview-13K, a training dataset that captures the intermediate reasoning processes inherent in academic paper reviews, addressing the fundamental challenges in Paper Review tasks from three dimensions: the scarcity of high-quality, structured review datasets and standardized evaluation frameworks. 

\subsection{DeepReview-13K}

\begin{table}[h]
\centering

\label{tab:dataset_statistics}
\resizebox{0.58\textwidth}{!}{
\begin{tabular}{@{}lcccc@{}}
\toprule
Dataset & Number & Tokens &  Rating & Accept Rate \\
\midrule
ICLR 2024 Train & 4131  & 10439 & 5.34 & 37.8\% \\
ICLR 2025 Train & 9247 & 10062 & 5.13 & 31.2\% \\
\cellcolor[HTML]{F9EBE3}{DeepReview-13K} & \cellcolor[HTML]{F9EBE3}{13378} & \cellcolor[HTML]{F9EBE3}{10178} & \cellcolor[HTML]{F9EBE3}{5.18} & \cellcolor[HTML]{F9EBE3}{33.24\%} \\
\midrule
ICLR 2024 Test & 652 & 10681 & 5.47 & 43.7\% \\
ICLR 2025 Test & 634 & 10241 &5.18 & 31.1\% \\
\cellcolor[HTML]{F9EBE3}{DeepReview-Bench} & \cellcolor[HTML]{F9EBE3}{1286} & \cellcolor[HTML]{F9EBE3}{10464} &\cellcolor[HTML]{F9EBE3}{5.33} & \cellcolor[HTML]{F9EBE3}{37.49\%} \\
\bottomrule
\end{tabular}}
\vspace{0.25cm}
\caption{Dataset Statistics. The table shows the average values of Tokens, Rating, and Accept Rate}
\label{tab:stat}
\end{table}
The statistics of this dataset are detailed in Table \ref{tab:stat}. We initially collected raw data from the OpenReview platform arXiv repository, gathering 18,976 paper submissions spanning two ICLR conference cycles (2024-2025)\footnote{Empty PDFs were filtered during conversion}. Using the MinerU tool \citep{wang2024mineruopensourcesolutionprecise}, we convert papers to parseable Markdown format, prioritizing \LaTeX{} source code when available from arXiv. For each paper, we assembled a review set $\mathbf{R}$ comprising three key components:  (1) textual assessments (Strengths, Weaknesses, and Questions), (2) interactive discussions from the rebuttal phase, and (3) standardized scores, including overall ratings ($\in [1,10]$) and fine-grained evaluations of Soundness, Presentation, and Contribution ($\in [1,4]$). Additionally, we collect meta-review texts and final ratings with acceptance decisions. The final DeepReview-13K dataset comprises 13,378 valid samples in Table \ref{tab:stat} as the foundation for constructing our review reasoning chain.



\subsection{DeepReview-Test}
To evaluate performance, we randomly sampled 10\% (1.2K) of the dataset to create DeepReview-Bench. Our evaluation framework assesses both quantitative scores and qualitative aspects of review generation through the following tasks:

\textbf{Quantitative Evaluation:}
1) Rating prediction: using MAE, MSE, accuracy, and F1 metrics
2) Paper quality ranking: measured by Spearman correlation
3) Pairwise paper selection (n=2): assessed through accuracy

\textbf{Qualitative Evaluation:}
While previous work \citep{tan2024peerreviewmultiturnlongcontext} relied on simple text similarity metrics (e.g., ROUGE \citep{lin-2004-rouge}, BLEU \citep{10.3115/1073083.1073135}), these metrics fail to capture specific review capabilities. Motivated by recent findings \citep{li2024llmasajudge}, we adopt the LLM-as-a-judge paradigm using Gemini-2.0-Flash-Thinking to conduct pairwise comparative evaluations of generated reviews. Detailed evaluation metrics are provided in Appendix \ref{appendix:eval}.


\section{Methodology}

Drawing inspiration from recent advances in complex reasoning methods \citep{xiang20252reasoningllmslearning,hao2024traininglargelanguagemodels}, we propose a deep-thinking evaluation framework that decomposes the review process into three key steps in Figure \ref{fig:main}: (1) novelty verification $z_1$: assessing research originality through literature review; (2) multi-dimension evaluation $z_2$: synthesizing insights from multiple expert perspectives; and (3) reliability verification $z_3$: examining internal consistency and logical coherence.

\subsection{Task Definition}

Formally, given an input paper $\mathbf{q}$, our goal is to generate a review pair $(\mathbf{s}, \mathbf{a})$, where $\mathbf{s}$ represents the qualitative assessment text (meta-review), we express the reasoning process as:
\begin{equation*}
\mathbf{q} \rightarrow z_1 \rightarrow z_2 \rightarrow z_3 \rightarrow (\mathbf{s}, \mathbf{a})
\end{equation*}

We formulate the review score generation as a marginalization over sequential reasoning chains:
\begin{equation}
p(\mathbf{a}|\mathbf{q}) \propto \int p(\mathbf{a}|z_{1:3}, \mathbf{q}) \prod_{t=1}^{3} p(z_t | z_{<t}, \mathbf{q})d\mathbf{Z}
\end{equation}
Here, the chain-of-thought term $\prod_{t=1}^{3} p(z_t | z_{<t}, \mathbf{q})$ explicitly models the sequential dependencies between reasoning steps, $\mathbf{Z}$ represents all possible intermediate state sequences $(s_1,\dots,s_n)$. This structured approach aims to enhance the reliability of the evaluation process. 

\subsection{Structured Reasoning Process}
\label{sec:4.2}
We present a comprehensive automated data construction pipeline, which is specifically designed to generate high-quality supervised fine-tuning datasets that capture complete reasoning paths, shown as $(z_1,z_2,z_3)$.

\textbf{Stage 1: Novelty Verification  ($z_1$).}
Our novelty verification framework consists of three key components: \textbf{question generation}, \textbf{paper analysis}, and \textbf{literature review}. Initially, based on the paper, we use the Qwen-2.5-72B-Instruct model \citep{qwen2025qwen25technicalreport} to generate three key research questions, focusing on research gaps, innovative directions, and methodological breakthroughs to capture domain-specific characteristics. Additionally, to ensure thorough understanding, we employ the Gemini-2.0-Flash-thinking model to conduct systematic paper analysis with a specifically designed system prompt (Figure \ref{fig:prompt2}) across research motivation, core ideas, technical approaches, and experimental design. Then, literature retrieval, comparison, and summary are built on OpenScholar \cite{asai2024openscholarsynthesizingscientificliterature} to address these research questions. Using Qwen-2.5-3B-Instruct with few-shot learning, we transform questions into search keywords to retrieve approximately 60 relevant papers via Semantic Scholar API. Subsequently, the ReRank model\footnote{\url{https://huggingface.co/OpenSciLM/OpenScholar_Reranker}} reorder retrieved papers and select the top 10 most relevant papers, and its internal QA model \footnote{\url{https://huggingface.co/OpenSciLM/Llama-3.1_OpenScholar-8B}} generates comprehensive reports as novelty analysis $z_1$, incorporating works cited in review $R$.



\textbf{Stage 2: Multi-dimension Review ($z_2$).} To provide constructive review, we transform author rebuttals into instructive suggestions while synthesizing multiple review $\mathbf{R}$ into comprehensive perspectives. Specifically, using Qwen-2.5-72B-Instruct, we develop a review reconstruction pipeline that analyzes each review in $\mathbf{R}$ with its corresponding author response, capturing experimental results, theoretical proofs, and implementation details from rebuttals to transform criticisms into concrete technical suggestions. The reconstruction process ($z_2$) follow three principles: (1) maintaining technical depth; (2) ensuring actionable feedback; (3) preserving professional tone and original citations. 

\textbf{Stage 3: Reliability Verification($z_3$).} In order to ensure assessment accuracy through systematic evidence analysis, we employ Gemini-2-Flash-thinking to conduct systematic evidence analysis through a four-stage verification chain:   \textit{methodology verification}, \textit{experimental verification}, and \textit{comprehensive analysis}. Each review comment requires supporting evidence from the paper and receives an assigned confidence level. Finally, we utilize Qwen to generate a new Meta-Review by integrating the original Meta-Review, reviewer comments, and verification outcomes. This step identifies key weaknesses while providing evidence-based analysis and constructive suggestions.

\begin{table*}[h]
\centering

\resizebox{0.995\textwidth}{!}{
\begin{tabular}{@{}ll|cccc|c|c|cccc|c|c@{}}
\toprule
 & & \multicolumn{6}{c|}{ICLR 2024}  & \multicolumn{6}{c}{ICLR 2025}  \\
\cmidrule(lr){3-6} \cmidrule(lr){7-7} \cmidrule(lr){8-8} \cmidrule(lr){9-12} \cmidrule(lr){13-13} \cmidrule(lr){14-14}
Method & Model & \multicolumn{4}{c|}{Score} & Ranking & Selection & \multicolumn{4}{c|}{Score} & Ranking & Selection \\
\cmidrule(lr){3-6} \cmidrule(lr){7-7} \cmidrule(lr){8-8} \cmidrule(lr){9-12} \cmidrule(lr){13-13} \cmidrule(lr){14-14}
 &  & R. MSE↓ & R. MAE↓ & D. Acc.$\uparrow$ & D. F1$\uparrow$ & R. Spearman$\uparrow$ & Pair. R. Acc$\uparrow$ & R. MSE↓ & R. MAE↓ & D. Acc.$\uparrow$ & D. F1$\uparrow$ & R. Spearman$\uparrow$ & Pair. R. Acc$\uparrow$ \\

\midrule

\multirow{4}{*}{Agent Review}
 & Claude-3-5-sonnet  & 2.8878 & 1.2715 & 0.4333 & 0.3937 & 0.1564 & 0.5526 & 2.8406 & 1.2989 & 0.2826 & 0.2541 & -0.0219 & 0.5432  \\
 & Gemini-2.0-Flash-Thinking          & 3.1943 & 1.3418 & 0.4400 & 0.4318 & -0.0252 & 0.5044 & 2.6186 & 1.2170 & 0.4242 & 0.4242 & 0.0968  & 0.5496 \\
 & DeepSeek-V3                 & 1.9479  &1.0735 &0.4105 & 0.3403 & 0.3542  & 0.6096 & 1.9951 & 1.1017 & 0.3140 & 0.2506 & 0.1197 & 0.5702\\
\midrule
\multirow{5}{*}{AI Scientist}
 & GPT-o1  & 4.3414 & 1.7294 & 0.4500 & 0.4424 & 0.2621 & 0.5881 & 4.3072 & 1.7917 & 0.4167 & 0.4157 & 0.2991 & 0.6318 \\
 & Claude-3-5-sonnet & 3.4447 & 1.5037 & 0.4787 & 0.4513 & 0.0366 & 0.5305  & 3.0992 & 1.3500 & 0.5579 & 0.4440 & -0.0219 & 0.5169 \\
 & Gemini-2.0-Flash-Thinking & 4.9297 & 1.8711 & 0.5743 & 0.5197 & 0.0745 & 0.5343 & 3.9232 & 1.6470 & 0.6139 & 0.4808 & 0.2565 & 0.6040\\
 & DeepSeek-V3  & 4.7337 & 1.7888 & 0.5600 & 0.5484 & 0.2310 & 0.5844 & 4.8006 & 1.8403 & 0.4059 & 0.3988 & 0.0778 & 0.5473\\
 & DeepSeek-R1  & 4.1648 & 1.6526 & 0.5248 & 0.4988 & 0.3256 & 0.6206 & 4.7719 & 1.8099 & 0.4259 & 0.4161 & 0.3237 & 0.6289\\

\midrule

\multirow{2}{*}{CycleReviewer}
 & 8B  & 2.8911 & 1.2371 & 0.6353 & 0.5528 & 0.2801 & 0.5993 & 2.4461 & 1.2063 & 0.6780 & 0.5586 & 0.2786 & 0.5960 \\
 & 70B & 2.4870 & 1.2514 & 0.6304 & 0.5696 & 0.3356 & 0.6160 & 2.4294 & 1.2128 & 0.6782 & 0.5737 & 0.2674 & 0.5928 \\

\midrule
 DeepReviewer & 14B  & \cellcolor[HTML]{F9EBE3}\textbf{1.3137} & \cellcolor[HTML]{F9EBE3}\textbf{0.9102} & \cellcolor[HTML]{F9EBE3}\textbf{0.6406} & \cellcolor[HTML]{F9EBE3}\textbf{0.6307} & \cellcolor[HTML]{F9EBE3}\textbf{0.3559} & \cellcolor[HTML]{F9EBE3}\textbf{0.6242} & \cellcolor[HTML]{F9EBE3}\textbf{1.3410} & \cellcolor[HTML]{F9EBE3}\textbf{0.9243} & \cellcolor[HTML]{F9EBE3}\textbf{0.6878} & \cellcolor[HTML]{F9EBE3}\textbf{0.6227} & \cellcolor[HTML]{F9EBE3}\textbf{0.4047} & \cellcolor[HTML]{F9EBE3}\textbf{0.6402}\\

\bottomrule
\end{tabular}}
\caption{\textbf{Performance comparison of reviewer models on DeepReview-13k datasets}. Notes: Metrics are grouped into Score (Rating MSE, Rating MAE, Decision Accuracy, Decision F1), Ranking (Rating Spearman), and Selection (Pairwise Rating Accuracy). Abbreviations: R.=Rating, MSE=Mean Squared Error, MAE=Mean Absolute Error, D. Acc.=Decision Accuracy, D. F1=Decision F1 score, Pair. R. Acc.=Pairwise Rating Accuracy.}
\label{tab:mresults}
\end{table*}

\textbf{Quality Control Mechanism.} To ensure the high quality of our synthetic DeepReview-13K dataset, we implemented a rigorous automated quality control process using Qwen-2.5-72B-Instruct. This process involves a multi-faceted approach to assess each generated sample for logical integrity and completeness. Specifically, Qwen-2.5-72B-Instruct was tasked with examining each sample for: (1) Logical Consistency: verifying that the reasoning chain ($z_1, z_2, z_3$) and the final evaluation $(\mathbf{s}, \mathbf{a})$ are logically coherent and non-contradictory; (2) Completeness: checking for any missing or empty fields within the structured data format, ensuring all components of the reasoning path and evaluation are present. Samples failing any of these checks, indicating logical inconsistencies, incompleteness, or failing to meet our quality standards, were automatically flagged and removed from the dataset.

\subsection{Model Training}
We train our model based on Phi-4 14B~\citep{abdin2024phi4technicalreport} using the DeepReview-13K dataset.  The training process was conducted on 8x H100 80G GPUs with DeepSpeed + ZeRO3 \citep{rajbhandari2020zeromemoryoptimizationstraining,10.1145/3394486.3406703} for optimization. Notably, we extended the context window to 256K using LongRoPE \citep{ding2024longrope}, with a 40K context window during training for full-parameter fine-tuning. Given memory constraints, samples exceeding the preset context length are randomly truncated. The model is trained for 23,500 steps with a batch size of 16 and a learning rate of 5e-6. 

\textbf{Inference Strategy.}
We divided each sample in the DeepReview-13K data into three modes using reasoning path cropping, as shown in Figure \ref{fig:main}(c), which allows for efficiency adjustments at test time based on varying requirements. The \textit{Fast} mode directly generates final evaluation results and comprehensive analysis reports $(\mathbf{s}, \mathbf{a})$, minimizing computational cost by bypassing intermediate reasoning steps. The \textit{Standard} mode executes core evaluation steps including $z_2$ and $z_3$, maintaining high efficiency while ensuring evaluation quality, making it appropriate for routine research assessment. The \textit{Best} mode implements the complete reasoning chain $(z_1,z_2,z_3)$, encompassing novelty verification, multi-dimension assessment, reliability verification, and comprehensive analysis generation. For novelty verification during inference, as in Stage 1 (Section \ref{sec:4.2}), we employ Semantic Scholar API and OpenScholar to ensure accurate assessment of research novelty and citation correctness through comprehensive literature review and analysis. All three modes share the same model architecture, differing only in their executed evaluation steps. This allows the trained DeepReview-14B model to execute different reasoning paths at inference time, controlled by input instructions.
\section{Experiments}

\subsection{Experimental setting}

\textbf{Baselines.} We consider two types of baselines: (1) Prompt-based baselines including AI Scientist \citep{chris2024the} and AgentReview \citep{jin-etal-2024-agentreview} implemented with various backbone models (GPT-o1-2024-12-17, Claude-3.5-sonnet-20241022, Gemini-2.0-Flash-Thinking-01-21, DeepSeek-V3, and DeepSeek-R1); (2) Fine-tuned baselines including CycleReviewer-8B and CycleReviewer-70B, both trained on ICLR 2024 review data. For inference, we use a temperature of 0.4 with maximum input and output lengths set to 100K and 16,384 tokens respectively to ensure complete text processing.
\begin{table*}[h]
\centering

\resizebox{0.995\textwidth}{!}{
\begin{tabular}{@{}ll|cccccc|ccc|ccc@{}}
\toprule
 \multicolumn{2}{c|}{} & \multicolumn{6}{c|}{Score} & \multicolumn{3}{c|}{Ranking} & \multicolumn{3}{c}{Pairwise Accuracy} \\
\cmidrule(lr){3-8} \cmidrule(lr){9-11} \cmidrule(lr){12-14}
Method & Model & S. MSE↓ & S. MAE↓ & P. MSE↓ & P. MAE↓ & C. MSE↓ & C. MAE↓ & S. Spearman$\uparrow$ & P. Spearman$\uparrow$ & C. Spearman$\uparrow$ & Pair. S. Acc$\uparrow$ & Pair. P. Acc$\uparrow$ & Pair. C. Acc$\uparrow$ \\

\toprule
\multicolumn{14}{l}{\textit{\textbf{ICLR 2024}}} \\
\midrule
\multirow{5}{*}{AI Scientist} & GPT-o1 & 0.4589 & 0.5336 & 0.5483 & 0.5983 & 0.7550 & 0.7147 & 0.1872 & 0.0723 & 0.1103 & 0.5797 & 0.5407 & 0.5621 \\
 & Claude-3-5-sonnet & 0.3052 & 0.4388 & 0.4745 & 0.5504 & 1.1420 & 0.8876 & 0.1692 & 0.0178 & 0.0275 & 0.6017 & 0.5440 & 0.5726 \\
 & Gemini-2.0-Flash-Thinking & 0.7233 & 0.6224 & 0.5264 & 0.5797 & 0.9036 & 0.7480 & 0.1050 & 0.1561 & 0.0274 & 0.5853 & 0.5929 & 0.5471 \\
 & DeepSeek-V3 & 0.8810 & 0.7718 & 0.7662 & 0.7145 & 1.6936 & 1.1400 & 0.2258 & 0.3189 & 0.1574 & 0.6028 & 0.6242 & 0.5933 \\
 & DeepSeek-R1 & 1.0540 & 0.8629 & 0.5356 & 0.5746 & 1.9564 & 1.2967 & 0.1664 & 0.2927 & 0.3009 & 0.6091 & 0.6315 & \cellcolor[HTML]{F9EBE3}\textbf{0.6517} \\
\midrule
\multirow{2}{*}{CycleReviewer} & 8B & 0.2516 & 0.3917 & 0.2356 & 0.3686 & 0.2507 & 0.3941 & 0.1990 & 0.3324 & 0.2593 & 0.5769 & 0.6103 & 0.5923 \\
 & 70B & 0.2375 & 0.3897 & 0.2414 & 0.3737 & 0.2657 & 0.4052 & 0.2320 & 0.3373 & 0.2354 & 0.5829 & 0.6230 & 0.5896 \\
\midrule
DeepReviewer & 14B  & \cellcolor[HTML]{F9EBE3}\textbf{0.1578} & \cellcolor[HTML]{F9EBE3}\textbf{0.3029} & \cellcolor[HTML]{F9EBE3}\textbf{0.1896} & \cellcolor[HTML]{F9EBE3}\textbf{0.3291} & \cellcolor[HTML]{F9EBE3}\textbf{0.2173} & \cellcolor[HTML]{F9EBE3}\textbf{0.3680} & \cellcolor[HTML]{F9EBE3}\textbf{0.3204} & \cellcolor[HTML]{F9EBE3}\textbf{0.3784} & \cellcolor[HTML]{F9EBE3}\textbf{0.3335} & \cellcolor[HTML]{F9EBE3}\textbf{0.6175} & \cellcolor[HTML]{F9EBE3}\textbf{0.6353} & 0.6208 \\

\toprule
\multicolumn{14}{l}{\textit{\textbf{ICLR 2025}}} \\

\toprule
\multirow{5}{*}{AI Scientist} & GPT-o1 & 0.4513 & 0.5500 & 0.4878 & 0.5750 & 0.6734 & 0.6802 & -0.0390 & -0.2837 & 0.1671 & 0.5541 & 0.5426 & 0.5966 \\
 & Claude-3-5-Sonnet & 0.4565 & 0.5279 & 0.5804 & 0.6346 & 0.8251 & 0.7628 & -0.0814 & -0.0790 & -0.0051 & 0.5543 & 0.5272 & 0.5454 \\
 & Gemini-2.0-Flash-Thinking & 0.4279 & 0.5219 & 0.6337 & 0.6114 & 0.5696 & 0.5876 & 0.3565 & 0.0593 & 0.2773 & \cellcolor[HTML]{F9EBE3}\textbf{0.6535} & 0.5499 & \cellcolor[HTML]{F9EBE3}\textbf{0.6321} \\
 & DeepSeek-V3 & 0.7999 & 0.7409 & 0.9120 & 0.7657 & 2.0180 & 1.2594 & 0.1926 & 0.0621 & -0.0677 & 0.6014 & 0.5683 & 0.5315 \\
 & DeepSeek-R1 & 0.8575 & 0.7636 & 0.4884 & 0.5586 & 2.1620 & 1.3750 & 0.3130 & 0.3133 & 0.3060 & 0.6289 & 0.5989 & 0.6268 \\
\midrule
\multirow{2}{*}{CycleReviewer} & 8B & 0.2617 & 0.3931 & 0.2880 & 0.4208 & 0.2667 & 0.4112 & 0.2377 & 0.2498 & 0.2511 & 0.5913 & 0.6074 & 0.5919 \\
 & 70B & 0.2588 & 0.3998 & 0.2562 & 0.3998 & 0.2601 & 0.4034 & 0.2320 & 0.2772 & 0.1905 & 0.5865 & 0.6051 & 0.5775 \\
\midrule
DeepReviewer & 14B & \cellcolor[HTML]{F9EBE3}\textbf{0.2239} & \cellcolor[HTML]{F9EBE3}\textbf{0.3650} & \cellcolor[HTML]{F9EBE3}\textbf{0.2178} & \cellcolor[HTML]{F9EBE3}\textbf{0.3662} & \cellcolor[HTML]{F9EBE3}\textbf{0.2632} & \cellcolor[HTML]{F9EBE3}\textbf{0.4095} & \cellcolor[HTML]{F9EBE3}\textbf{0.3810} & \cellcolor[HTML]{F9EBE3}\textbf{0.3698} & \cellcolor[HTML]{F9EBE3}\textbf{0.3239} & 0.6057 & \cellcolor[HTML]{F9EBE3}\textbf{0.6380} & 0.6222 \\

\toprule
\end{tabular}}
\caption{\textbf{Performance comparison of reviewer models on fine-grained evaluation dimensions}. This table presents the performance across three key assessment aspects: Soundness (S.), Presentation (P.), and Contribution (C.) on ICLR 2024 and 2025 conferences.}
\label{tab:detailed_results_combined}
\end{table*}
\subsection{Main Results}

Test results are shown in Table \ref{tab:mresults}. Compared with prompt-based baselines, DeepReviewer reduces Rating MSE by an average of 65.83$\%$ and improves Decision Accuracy by an average of 15.2$\%$ points from AI Scientist. When compared to strong finetuned baseline CycleReviewer-70B, DeepReviewer represents reductions of 44.80\% for Rating MSE. For the critical accept/reject decision task, DeepReviewer achieves 64.06\% decision accuracy and 0.6307 F1 score on ICLR 2024, substantially surpassing all baselines. Notably, DeepReviewer with 14B parameters outperforms significantly larger models including CycleReviewer-70B (70B parameters) and other closed-source LLMs, demonstrating that DeepReviewer provides more reliable paper assessment than other approaches. 


DeepReviewer achieves the highest Rating Spearman correlations of 0.3559 and 0.4047 on ICLR 2024 and ICLR 2025 respectively, improving upon CycleReviewer-70B by 6.04$\%$ and AI Scientist (DeepSeek-R1) by 25.02$\%$. In the paper selection task, It demonstrates superior discrimination ability with pairwise accuracies of 0.62 and 0.64 on ICLR 2024 and ICLR 2025 respectively.



\begin{table*}[h]
\centering

\resizebox{0.995\textwidth}{!}{
\begin{tabular}{lcccccccc|ccc}
\toprule
\multirow{1}{*}{\textbf{Baselines}} & \multicolumn{2}{c}{\textbf{Constructive Value}} & \multicolumn{2}{c}{\textbf{Analytical Depth}} & \multicolumn{2}{c}{\textbf{Plausibility}} & \multicolumn{2}{c}{\textbf{Technical Accuracy}} & \multicolumn{2}{c}{\textbf{Overall Judgment}} \\
\cmidrule(lr){2-3} \cmidrule(lr){4-5} \cmidrule(lr){6-7} \cmidrule(lr){8-9} \cmidrule(lr){10-11}
 \textbf{DeepReviewer 14B vs.}& \textbf{Win(\%)$\uparrow$} & \textbf{Lose(\%)} & \textbf{Win(\%)$\uparrow$} & \textbf{Lose(\%)} & \textbf{Win(\%)$\uparrow$} & \textbf{Lose(\%)} & \textbf{Win(\%)$\uparrow$} & \textbf{Lose(\%)} & \textbf{Win(\%)$\uparrow$} & \textbf{Lose(\%)} \\
\midrule
\textbf{ICLR 2024} & & & & & & & & & & \\

\textit{AI Scientist} GPT-o1 & \cellcolor[HTML]{F9EBE3}\textbf{89.80} & 6.67 & \cellcolor[HTML]{F9EBE3}\textbf{87.67} & 6.67 & \cellcolor[HTML]{F9EBE3}\textbf{51.69} & 3.53 & \cellcolor[HTML]{F9EBE3}\textbf{25.12} & 11.67 & \cellcolor[HTML]{F9EBE3}\textbf{88.21} & 6.63 \\
\textit{AI Scientist} Claude-3.5-Sonnet & \cellcolor[HTML]{F9EBE3}\textbf{96.88} & 3.12 & \cellcolor[HTML]{F9EBE3}\textbf{97.92} & 2.08 & \cellcolor[HTML]{F9EBE3}\textbf{80.21} & 4.17 & \cellcolor[HTML]{F9EBE3}\textbf{77.08} & 2.08 & \cellcolor[HTML]{F9EBE3}\textbf{95.74} & 4.26 \\
\textit{AI Scientist} Gemini-2.0-Flash-Thinking & \cellcolor[HTML]{F9EBE3}\textbf{53.47} & 17.82 & \cellcolor[HTML]{F9EBE3}\textbf{53.47} & 20.79 & \cellcolor[HTML]{F9EBE3}\textbf{24.75} & 10.89 & 18.81 & \cellcolor[HTML]{F9EBE3}\textbf{20.79} & \cellcolor[HTML]{F9EBE3}\textbf{59.41} & 25.74 \\
\textit{AI Scientist} DeepSeek-V3 & \cellcolor[HTML]{F9EBE3}\textbf{96.04} & 1.98 & \cellcolor[HTML]{F9EBE3}\textbf{99.01} & 0.00 & \cellcolor[HTML]{F9EBE3}\textbf{72.28} & 0.99 & \cellcolor[HTML]{F9EBE3}\textbf{67.33} & 4.95 & \cellcolor[HTML]{F9EBE3}\textbf{96.22} & 0.00 \\
\textit{AI Scientist} DeepSeek-R1 & \cellcolor[HTML]{F9EBE3}\textbf{89.22} & 7.84 & \cellcolor[HTML]{F9EBE3}\textbf{74.51} & 13.73 & \cellcolor[HTML]{F9EBE3}\textbf{45.10} & 5.88 & \cellcolor[HTML]{F9EBE3}\textbf{26.47} & 18.63 & \cellcolor[HTML]{F9EBE3}\textbf{80.20} & 16.83 \\

\textit{AgentReview} Claude-3.5-Sonnet & \cellcolor[HTML]{F9EBE3}\textbf{96.84} & 1.05 & \cellcolor[HTML]{F9EBE3}\textbf{98.94} & 0.00 & \cellcolor[HTML]{F9EBE3}\textbf{90.43} & 0.00 & \cellcolor[HTML]{F9EBE3}\textbf{77.08} & 0.00 & \cellcolor[HTML]{F9EBE3}\textbf{98.90} & 0.00 \\
\textit{AgentReview} Gemini-2.0-Flash-Thinking & \cellcolor[HTML]{F9EBE3}\textbf{98.00} & 1.00 & \cellcolor[HTML]{F9EBE3}\textbf{95.11} & 1.00 & \cellcolor[HTML]{F9EBE3}\textbf{81.64} & 0.01 & \cellcolor[HTML]{F9EBE3}\textbf{65.00} & 3.00 & \cellcolor[HTML]{F9EBE3}\textbf{96.74} & 1.00 \\
\textit{AgentReview} GPT-4o & \cellcolor[HTML]{F9EBE3}\textbf{99.02} & 0.99 & \cellcolor[HTML]{F9EBE3}\textbf{99.01} & 0.99 & \cellcolor[HTML]{F9EBE3}\textbf{95.05} & 0.99 & \cellcolor[HTML]{F9EBE3}\textbf{61.76} & 4.90 & \cellcolor[HTML]{F9EBE3}\textbf{98.15} & 1.00 \\
\textit{CycleReviewer} 8B & \cellcolor[HTML]{F9EBE3}\textbf{97.30} & 1.80 & \cellcolor[HTML]{F9EBE3}\textbf{98.20} & 0.91 & \cellcolor[HTML]{F9EBE3}\textbf{90.92} & 0.91 & \cellcolor[HTML]{F9EBE3}\textbf{87.50} & 0.00 & \cellcolor[HTML]{F9EBE3}\textbf{96.09} & 0.91 \\
\textit{CycleReviewer} 70B & \cellcolor[HTML]{F9EBE3}\textbf{98.33} & 1.11 & \cellcolor[HTML]{F9EBE3}\textbf{98.89} & 0.01 & \cellcolor[HTML]{F9EBE3}\textbf{92.78} & 0.01 & \cellcolor[HTML]{F9EBE3}\textbf{79.44} & 0.01 & \cellcolor[HTML]{F9EBE3}\textbf{98.33} & 1.11 \\
\bottomrule
\textbf{ICLR 2025} & & & & & & & & & & \\
\textit{AI Scientist} GPT-o1 & \cellcolor[HTML]{F9EBE3}\textbf{91.67} & 8.33 & \cellcolor[HTML]{F9EBE3}\textbf{89.58} & 8.33 & \cellcolor[HTML]{F9EBE3}\textbf{60.42} & 4.17 & \cellcolor[HTML]{F9EBE3}\textbf{37.50} & 8.33 & \cellcolor[HTML]{F9EBE3}\textbf{91.67} & 8.33 \\
\textit{AI Scientist} Claude-3.5-Sonnet & \cellcolor[HTML]{F9EBE3}\textbf{97.87} & 1.06 & \cellcolor[HTML]{F9EBE3}\textbf{100.00} & 0.00 & \cellcolor[HTML]{F9EBE3}\textbf{92.55} & 1.06 & \cellcolor[HTML]{F9EBE3}\textbf{65.96} & 0.00 & \cellcolor[HTML]{F9EBE3}\textbf{98.94} & 1.06 \\
\textit{AI Scientist} Gemini-2.0-Flash-Thinking & \cellcolor[HTML]{F9EBE3}\textbf{52.43} & 18.45 & \cellcolor[HTML]{F9EBE3}\textbf{52.43} & 23.30 & \cellcolor[HTML]{F9EBE3}\textbf{33.98} & 7.77 & 19.42 & \cellcolor[HTML]{F9EBE3}\textbf{20.39} & \cellcolor[HTML]{F9EBE3}\textbf{59.41} & 24.75 \\

\textit{AI Scientist} DeepSeek-V3 & \cellcolor[HTML]{F9EBE3}\textbf{96.04} & 2.97 & \cellcolor[HTML]{F9EBE3}\textbf{97.03} & 1.98 & \cellcolor[HTML]{F9EBE3}\textbf{75.25} & 2.97 & \cellcolor[HTML]{F9EBE3}\textbf{63.37} & 3.96 & \cellcolor[HTML]{F9EBE3}\textbf{97.03} & 2.97 \\
\textit{AI Scientist} DeepSeek-R1 & \cellcolor[HTML]{F9EBE3}\textbf{89.29} & 6.25 & \cellcolor[HTML]{F9EBE3}\textbf{81.25} & 10.71 & \cellcolor[HTML]{F9EBE3}\textbf{51.79} & 5.36 & \cellcolor[HTML]{F9EBE3}\textbf{26.79} & 18.75 & \cellcolor[HTML]{F9EBE3}\textbf{87.39} & 9.01 \\

\textit{AgentReview} Claude-3.5-Sonnet & \cellcolor[HTML]{F9EBE3}\textbf{95.74} & 1.06 & \cellcolor[HTML]{F9EBE3}\textbf{97.85} & 2.15 & \cellcolor[HTML]{F9EBE3}\textbf{90.32} & 2.15 & \cellcolor[HTML]{F9EBE3}\textbf{74.74} & 1.05 & \cellcolor[HTML]{F9EBE3}\textbf{97.83} & 2.17 \\
\textit{AgentReview} Gemini-2.0-Flash-Thinking & \cellcolor[HTML]{F9EBE3}\textbf{92.16} & 1.96 & \cellcolor[HTML]{F9EBE3}\textbf{93.08} & 3.00 & \cellcolor[HTML]{F9EBE3}\textbf{78.20} & 0.65 & \cellcolor[HTML]{F9EBE3}\textbf{61.76} & 4.90 & \cellcolor[HTML]{F9EBE3}\textbf{92.16} & 4.90 \\
\textit{AgentReview} GPT-4o & \cellcolor[HTML]{F9EBE3}\textbf{95.28} & 2.09 & \cellcolor[HTML]{F9EBE3}\textbf{95.37} & 1.40 & \cellcolor[HTML]{F9EBE3}\textbf{92.10} & 0.85 & \cellcolor[HTML]{F9EBE3}\textbf{65.03} & 5.47 & \cellcolor[HTML]{F9EBE3}\textbf{94.15} & 2.39 \\

\textit{CycleReviewer} 8B & \cellcolor[HTML]{F9EBE3}\textbf{98.45} & 1.55 & \cellcolor[HTML]{F9EBE3}\textbf{98.24} & 1.89 & \cellcolor[HTML]{F9EBE3}\textbf{86.37} & 0.77 & \cellcolor[HTML]{F9EBE3}\textbf{86.36} & 2.27 & \cellcolor[HTML]{F9EBE3}\textbf{98.45} & 1.55 \\
\textit{CycleReviewer} 70B & \cellcolor[HTML]{F9EBE3}\textbf{96.17} & 1.64 & \cellcolor[HTML]{F9EBE3}\textbf{96.17} & 2.19 & \cellcolor[HTML]{F9EBE3}\textbf{86.34} & 1.64 & \cellcolor[HTML]{F9EBE3}\textbf{72.68} & 3.28 & \cellcolor[HTML]{F9EBE3}\textbf{96.72} & 1.64 \\
\bottomrule
\end{tabular}}
\caption{Direct comparison of DeepReviewer with the baselines on general alignment tasks. Win indicates that Gemini-2.0-Flash-Thinking assesses DeepReviewer's response as superior compared to the baseline. Cells marked in light gray suggest baseline the winner.}

\label{tab:win_rate_comparison_deepreviewer}
\end{table*}

Table~\ref{tab:detailed_results_combined} presents a detailed analysis across three critical dimensions: Soundness, Presentation, and Contribution. Particularly for Soundness assessment on ICLR 2024, DeepReviewer-14B achieves an MSE of 0.1578 and MAE of 0.3029, representing improvements of 33.58$\%$ and 22.09$\%$ over CycleReviewer-70B. While DeepReviewer shows marginally lower performance than AI Scientist (Gemini-2.0-Flash-Thinking) in Contribution and Soundness accuracy, it maintains a balanced and strong performance across all dimensions. 

We observe a strong correlation between fine-grained assessment capability and overall rating performance. Models that excel in dimension-specific evaluations, such as DeepReviewer and Claude-3.5-Sonnet, consistently demonstrate superior performance in overall ratings. This pattern validates the effectiveness of our multi-stage reasoning chain design, particularly the necessity of multi-facet evaluation in our framework.

\subsection{Review Text Quality}

Table~\ref{tab:win_rate_comparison_deepreviewer} shows that DeepReviewer's overwhelming advantages across all evaluation dimensions. Interestingly, in the comparison with AI Scientist (Gemini-2.0-Flash-Thinking), despite being used as the judge, Gemini assessed most reviews in favor of DeepReviewer (winning 53.47\% in constructive value and analytical depth), with only two dimensions showing preference for its own reviews (20.79\% in technical accuracy). This self-critical evaluation further validates the objectivity of our assessment framework. In terms of overall judgment, DeepReviewer achieves remarkable win rates of 88.21\% against AI Scientist (GPT-o1) and 98.15\% against AgentReview (GPT-4o) on ICLR 2024.

The advantages are most prominent in constructive value and analytical depth. When compared with AgentReview (GPT-4o), DeepReviewer achieves win rates of 99.02\% and 99.01\% respectively, indicating that our Deep review with Thinking framework generates more insightful analysis and actionable suggestions. These qualitative assessments corroborate our quantitative findings, further validating the effectiveness of the multi-stage reasoning approach in our framework.
\begin{figure*}[t]
    \centering
    \includegraphics[width=\textwidth]{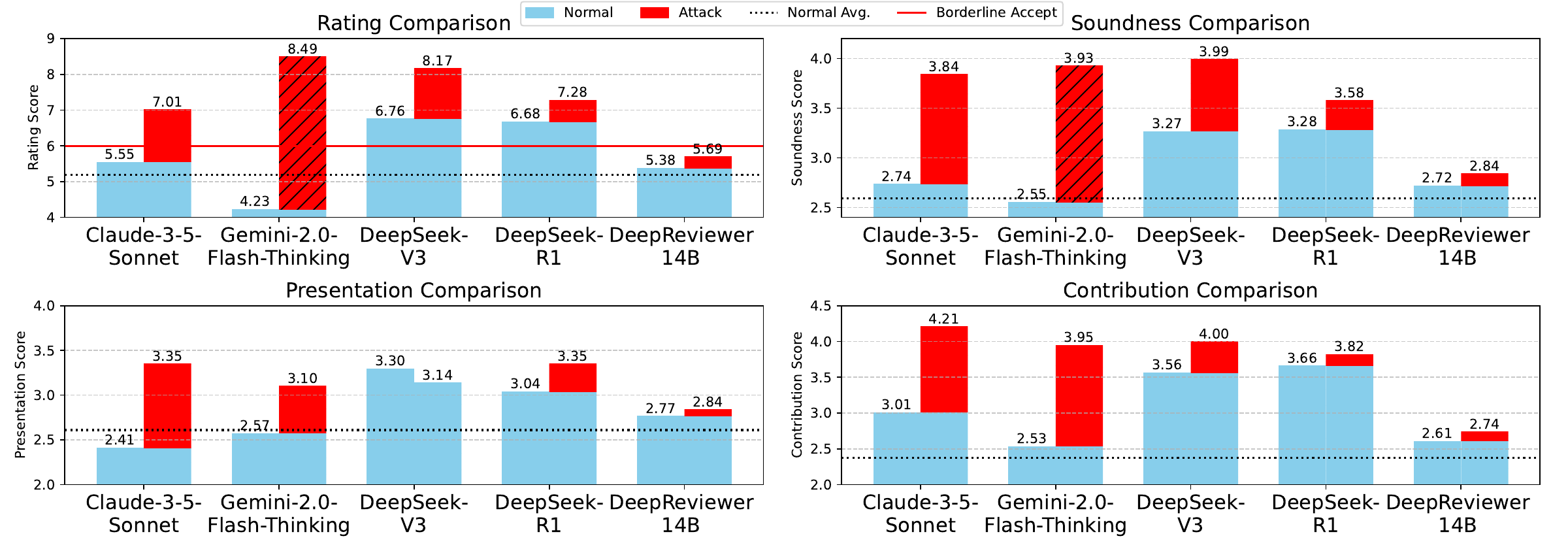}
    \vspace{-0.5cm}
    \caption{Demonstrates the scoring comparison of AI Scientist and DeepReviewer 14B models under normal and attack scenarios. The DeepReviewer model shows the smallest increase in scores (the growth of red bars relative to blue bars in the graph) when under attack, indicating its stronger robustness.}
    \vspace{-0.3cm}
    \label{fig:reject}
\end{figure*}

\begin{figure*}[t]
    \centering
    \includegraphics[width=\textwidth]{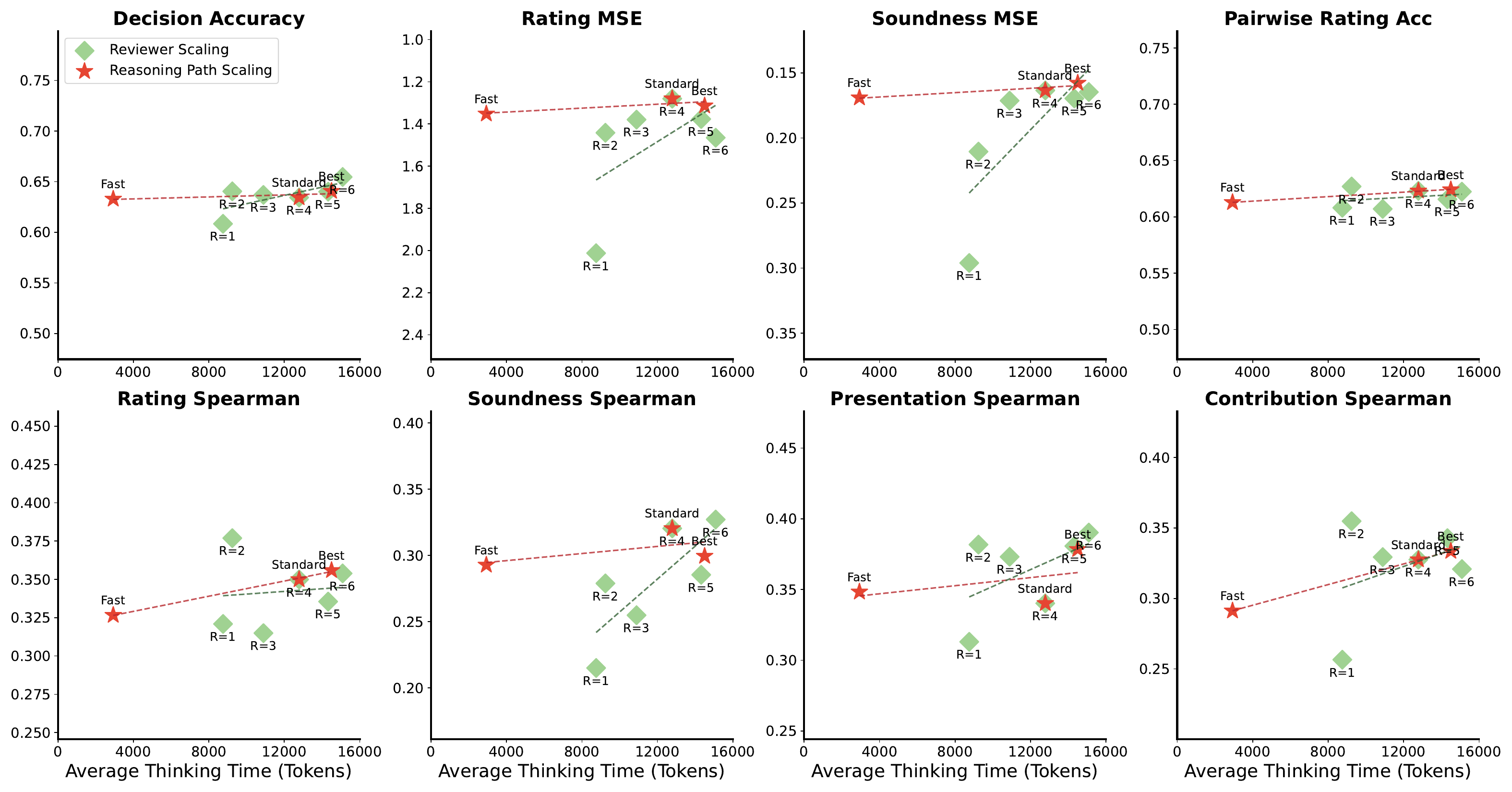}
    \vspace{-0.45cm}
    \caption{The performance of the DeepReviewer model in the Test-Time Scaling experiment. The x-axis represents the number of Tokens generated during model inference, and the y-axis represents different evaluation metrics. The green and red dashed lines are linear regression fitting curves for Reasoning Path Scaling and Reviewer Scaling scaling methods, respectively.}
    \vspace{-0.3cm}
    \label{fig:test}
\end{figure*}
\subsection{Defend Attacks Analysis}

We evaluate DeepReviewer's robustness against adversarial attacks \citep{ye2024we} by inserting malicious instructions into input papers. Figure~\ref{fig:reject} illustrates the rating comparison under normal and attack scenarios across different dimensions. Though not specifically trained with any adversarial samples, The DeepReviewer model demonstrates superior robustness compared to baseline systems. Under attack, the overall rating increase for DeepReviewer is merely 0.31 points (from 5.38 to 5.69), while other systems show substantial vulnerability, for example, Gemini-2.0-Flash-Thinking exhibits a dramatic increase of 4.26 points (from 4.23 to 8.49) and DeepSeek-V3 shows a 1.41 increase (from 6.76 to 8.17). This pattern held across fine-grained dimensions: for instance, Soundness scores for DeepReviewer increased by only 0.12 points, compared to larger increases for Claude-3.5-Sonnet (1.10) and Gemini-2.0-Flash-Thinking (1.38). We attribute this robustness to DeepReviewer's multi-stage reasoning framework, which, unlike direct input-output models, including content understanding, novelty verification, and reliability checks. It enabling a focus on intrinsic paper quality despite malicious prompts. However, the slight score increases under attack suggest room for improvement, we suggest that incorporating adversarial samples during training.



\subsection{Test-Time Scalability Study}

DeepReviewer model features unique test-time scaling capabilities through two mechanisms, both controllable via input instructions: Reasoning Path Scaling and Reviewer Scaling. Reasoning Path Scaling offers three inference modes—Fast, Standard, and Best—with progressively deeper reasoning and corresponding output token lengths of approximately 3,000, 8,000, and 14,500 tokens, respectively. Complementing this, Reviewer Scaling, employed within Standard mode, adjusts the number of simulated reviewers from R=1 to R=6. It enabling the synthesis of multi-perspective evaluations through simulated reviewer collaboration. Both scaling mechanisms inherently extend the model's evaluation process: Reasoning Path Scaling by increasing analytical depth, and Reviewer Scaling by emulating collaborative review.

\paragraph{Performance Analysis.} Figure~\ref{fig:test} illustrates significant performance enhancements as inference computation increases. In Reasoning Path Scaling (red stars), switching from Fast to Best mode results in steady improvements across all metrics, with the Rating Spearman correlation increasing by 8.97$\%$ (from 0.326 to 0.355). Reviewer Scaling (green diamonds) presents more diverse patterns across various tasks. In scoring tasks (Decision Accuracy, Rating MSE, Soundness MSE), consistent performance gains are observed with additional reviewers, indicating that score aggregation is enhanced by multiple viewpoints. The performance variability in Reviewer Scaling, especially when $R \neq 4$, likely arises from the model's training distribution being focused around four reviewers. Despite some variability, both scaling methods show positive trends (see regression lines), indicating our framework effectively uses more computational resources. The benefits vary by metric: scoring tasks improve most, followed by ranking, then selection. This suggests that multi-stage reasoning excels in complex paper evaluations, while simpler comparisons (e.g., choosing between two papers) gain less from added reasoning.

Furthermore, we observe that DeepReviewer's Fast mode, with only half the output tokens (3000), outperformed the CycleReviewer model (6000 output tokens) across various metrics (See Table \ref{tab:detailed_results_combined}), including Decision Accuracy, Rating MSE, and fine-grained Spearman correlations for Soundness, Presentation, and Contribution. Despite its simplified reasoning path, Fast mode retains core evaluation logic, such as identifying key paper content and critical flaws. We show that DeepReviewer utilizes each token more effectively, focusing on the most crucial information and achieving high performance with fewer output tokens.

Despite these variations, both scaling approaches demonstrate positive trends across metrics, validating that increased computational investment -- whether through more sophisticated inference modes or additional simulated reviewers -- enhances the model's paper assessment capabilities. 

\section{Conclusions}
We presented DeepReviewer, a novel framework for research paper evaluation aimed at enhancing the reliability of LLMs in paper reviews. DeepReviewer achieves adaptable reasoning depth through Test-Time Scaling to meet diverse needs. Our contributions are threefold: (1) the creation of DeepReview-13K, a detailedly annotated dataset that facilitates training for systematic and deep paper evaluation; (2) the training of the DeepReviewer model; and (3) comprehensive validation of DeepReviewer's superiority in both objective and subjective assessments. Notably, we explored and demonstrated effective Test-Time Scaling through Reasoning Path and Reviewer Scaling strategies.

\section*{Limitations}
Firstly, our approach relies on a synthetic dataset, DeepReview-13K, constructed through an automated pipeline. Although meticulously designed to mimic expert review processes and incorporating quality control mechanisms, this synthetic data may not fully capture the complexities and nuances of genuine human paper review. We have strived to mitigate this by leveraging real-world review data from ICLR conferences and incorporating structured reasoning annotations, but the inherent limitations of synthetic data persist. Secondly, while DeepReviewer offers Test-Time Scaling for efficiency, the "Best" mode, which employs the complete reasoning chain and external knowledge retrieval, can be computationally intensive. We address this by providing "Fast" and "Standard" modes, allowing for a trade-off between thoroughness and computational cost, catering to diverse application needs. Furthermore, while we have shown robustness against adversarial attacks, complete immunity is not yet achieved, indicating a need for ongoing research into enhancing security and reliability. Despite these limitations, DeepReviewer represents a significant step towards more reliable and robust LLM-based paper review systems, and our exploration of robust structured reasoning opens avenues for future research.

\section*{Ethical Considerations}

The development of DeepReviewer, while holding significant promise for enhancing the efficiency and potentially the quality of scholarly paper review, inherently carries ethical considerations that demand careful attention. We recognize that automating aspects of the peer review process introduces risks of bias amplification, deskilling of human reviewers, and a potential erosion of transparency and accountability. Specifically, DeepReviewer, like any LLM, could inadvertently perpetuate or even amplify existing biases present in the training data or encoded within its architecture. This could lead to systematic disadvantages for research from underrepresented groups, novel or unconventional methodologies, or topics perceived as less mainstream, even if the DeepReview-13K dataset was synthetically generated to be representative and fair. Furthermore, over-reliance on automated review assistance might diminish the critical thinking skills of human reviewers, potentially leading to a deskilling effect over time and a dependence on AI-driven assessments without sufficient human oversight.

To proactively address these ethical concerns and mitigate potential harms, we have implemented a multi-faceted approach throughout DeepReviewer's development and deployment. Firstly, while our training data is synthetic, we have rigorously designed the DeepReview-13K dataset and its generation pipeline to explicitly model expert reviewer reasoning and incorporate diverse perspectives, aiming to minimize the introduction of unintended biases. Secondly, we emphasize that DeepReviewer is intended as a decision support tool, designed to augment, not replace, human expertise. We strongly advocate for a human-in-the-loop approach, where DeepReviewer's outputs are critically evaluated and contextualized by expert reviewers. To ensure transparency, we are releasing DeepReviewer as an open-source resource, allowing for community scrutiny of its code, architecture, and potential biases. Alongside the code release, we will provide comprehensive user guidelines and best practices that explicitly caution against over-reliance on automated outputs and emphasize the importance of human oversight and critical assessment. Furthermore, our open-source licensing, while permissive, mandates that users disclose their institutional affiliation, personal information, and intended use case upon downloading DeepReviewer. This measure aims to foster accountability and enable a feedback loop, allowing us to monitor real-world applications, gather user feedback, and iteratively improve the model and its ethical safeguards. We also commit to ongoing bias auditing and benchmarking of DeepReviewer across diverse datasets and review scenarios, continually evaluating its performance and identifying areas for refinement. We believe these proactive measures, combined with ongoing community engagement and responsible user practices, are crucial to harnessing the benefits of DeepReviewer while minimizing its potential for harm and ensuring its ethical and beneficial application within the scientific peer review process.

\bibliographystyle{abbrvnat}
\bibliography{custom}

\begin{thebibliography}{72}
\providecommand{\natexlab}[1]{#1}
\providecommand{\url}[1]{\texttt{#1}}
\expandafter\ifx\csname urlstyle\endcsname\relax
  \providecommand{\doi}[1]{doi: #1}\else
  \providecommand{\doi}{doi: \begingroup \urlstyle{rm}\Url}\fi

\bibitem[Abdin et~al.(2024)Abdin, Aneja, Behl, Bubeck, Eldan, Gunasekar, Harrison, Hewett, Javaheripi, Kauffmann, Lee, Lee, Li, Liu, Mendes, Nguyen, Price, de~Rosa, Saarikivi, Salim, Shah, Wang, Ward, Wu, Yu, Zhang, and Zhang]{abdin2024phi4technicalreport}
M.~Abdin, J.~Aneja, H.~Behl, S.~Bubeck, R.~Eldan, S.~Gunasekar, M.~Harrison, R.~J. Hewett, M.~Javaheripi, P.~Kauffmann, J.~R. Lee, Y.~T. Lee, Y.~Li, W.~Liu, C.~C.~T. Mendes, A.~Nguyen, E.~Price, G.~de~Rosa, O.~Saarikivi, A.~Salim, S.~Shah, X.~Wang, R.~Ward, Y.~Wu, D.~Yu, C.~Zhang, and Y.~Zhang.
\newblock Phi-4 technical report, 2024.
\newblock URL \url{https://arxiv.org/abs/2412.08905}.

\bibitem[Achiam et~al.(2023)Achiam, Adler, Agarwal, Ahmad, Akkaya, Aleman, Almeida, Altenschmidt, Altman, Anadkat, et~al.]{achiam2023gpt}
J.~Achiam, S.~Adler, S.~Agarwal, L.~Ahmad, I.~Akkaya, F.~L. Aleman, D.~Almeida, J.~Altenschmidt, S.~Altman, S.~Anadkat, et~al.
\newblock Gpt-4 technical report.
\newblock \emph{arXiv preprint arXiv:2303.08774}, 2023.

\bibitem[AI(2025)]{Aider}
A.~AI.
\newblock Aider is ai pair programming in your terminal.
\newblock \url{https://github.com/Aider-AI/aider}, 2025.

\bibitem[Aky{\"u}rek et~al.(2022)Aky{\"u}rek, Schuurmans, Andreas, Ma, and Zhou]{akyurek2022learning}
E.~Aky{\"u}rek, D.~Schuurmans, J.~Andreas, T.~Ma, and D.~Zhou.
\newblock What learning algorithm is in-context learning? investigations with linear models.
\newblock \emph{arXiv preprint arXiv:2211.15661}, 2022.

\bibitem[Alberts et~al.(2008)Alberts, Hanson, and Kelner]{alberts2008reviewing}
B.~Alberts, B.~Hanson, and K.~L. Kelner.
\newblock Reviewing peer review, 2008.

\bibitem[Asai et~al.(2024)Asai, He, Shao, Shi, Singh, Chang, Lo, Soldaini, Feldman, D'arcy, Wadden, Latzke, Tian, Ji, Liu, Tong, Wu, Xiong, Zettlemoyer, Neubig, Weld, Downey, tau Yih, Koh, and Hajishirzi]{asai2024openscholarsynthesizingscientificliterature}
A.~Asai, J.~He, R.~Shao, W.~Shi, A.~Singh, J.~C. Chang, K.~Lo, L.~Soldaini, S.~Feldman, M.~D'arcy, D.~Wadden, M.~Latzke, M.~Tian, P.~Ji, S.~Liu, H.~Tong, B.~Wu, Y.~Xiong, L.~Zettlemoyer, G.~Neubig, D.~Weld, D.~Downey, W.~tau Yih, P.~W. Koh, and H.~Hajishirzi.
\newblock Openscholar: Synthesizing scientific literature with retrieval-augmented lms, 2024.
\newblock URL \url{https://arxiv.org/abs/2411.14199}.

\bibitem[Bai et~al.(2023)Bai, Bai, Chu, Cui, Dang, Deng, Fan, Ge, Han, Huang, et~al.]{bai2023qwen}
J.~Bai, S.~Bai, Y.~Chu, Z.~Cui, K.~Dang, X.~Deng, Y.~Fan, W.~Ge, Y.~Han, F.~Huang, et~al.
\newblock Qwen technical report.
\newblock \emph{arXiv preprint arXiv:2309.16609}, 2023.

\bibitem[Blog(2024)]{iclr2025}
I.~Blog.
\newblock Iclr 2025: Assisting reviewers.
\newblock \url{https://blog.iclr.cc/2024/10/09/iclr2025-assisting-reviewers/}, 2024.
\newblock Accessed: 2024-10-09.

\bibitem[Chris et~al.(2024)Chris, Cong, Robert, Jakob, Jeff, and David]{chris2024the}
L.~Chris, L.~Cong, L.~Robert, Tjarko, F.~Jakob, C.~Jeff, and H.~David.
\newblock The ai scientist: Towards fully automated open-ended scientific discovery.
\newblock \emph{arXiv preprint arXiv:2408.06292v3}, 2024.
\newblock URL \url{https://www.arxiv.org/abs/2408.06292v3}.

\bibitem[Daniil et~al.(2023)Daniil, Robert, and Gabe]{daniil2023emergent}
B.~Daniil, A., M.~Robert, and G.~Gabe.
\newblock Emergent autonomous scientific research capabilities of large language models.
\newblock \emph{arXiv preprint arXiv:2304.05332v1}, 2023.
\newblock URL \url{https://www.arxiv.org/abs/2304.05332v1}.

\bibitem[D'Arcy et~al.(2024)D'Arcy, Hope, Birnbaum, and Downey]{d2024marg}
M.~D'Arcy, T.~Hope, L.~Birnbaum, and D.~Downey.
\newblock Marg: Multi-agent review generation for scientific papers.
\newblock \emph{arXiv preprint arXiv:2401.04259}, 2024.

\bibitem[Ding et~al.(2024)Ding, Zhang, Zhang, Xu, Shang, Xu, Yang, and Yang]{ding2024longrope}
Y.~Ding, L.~L. Zhang, C.~Zhang, Y.~Xu, N.~Shang, J.~Xu, F.~Yang, and M.~Yang.
\newblock Longro{PE}: Extending {LLM} context window beyond 2 million tokens.
\newblock In \emph{Forty-first International Conference on Machine Learning}, 2024.
\newblock URL \url{https://openreview.net/forum?id=ONOtpXLqqw}.

\bibitem[Drori and Te'eni(2024)]{drori2024human}
I.~Drori and D.~Te'eni.
\newblock Human-in-the-loop ai reviewing: Feasibility, opportunities, and risks.
\newblock \emph{Journal of the Association for Information Systems}, 25\penalty0 (1):\penalty0 98--109, 2024.

\bibitem[Du et~al.(2024)Du, Wang, Zhao, Deng, Liu, Lou, Zou, Narayanan~Venkit, Zhang, Srinath, Zhang, Gupta, Li, Li, Wang, Liu, Liu, Gao, Xia, Xing, Jiayang, Wang, Su, Shah, Guo, Gu, Li, Wei, Wang, Cheng, Ranathunga, Fang, Fu, Liu, Huang, Blanco, Cao, Zhang, Yu, and Yin]{du2024llms}
J.~Du, Y.~Wang, W.~Zhao, Z.~Deng, S.~Liu, R.~Lou, H.~P. Zou, P.~Narayanan~Venkit, N.~Zhang, M.~Srinath, H.~R. Zhang, V.~Gupta, Y.~Li, T.~Li, F.~Wang, Q.~Liu, T.~Liu, P.~Gao, C.~Xia, C.~Xing, C.~Jiayang, Z.~Wang, Y.~Su, R.~S. Shah, R.~Guo, J.~Gu, H.~Li, K.~Wei, Z.~Wang, L.~Cheng, S.~Ranathunga, M.~Fang, J.~Fu, F.~Liu, R.~Huang, E.~Blanco, Y.~Cao, R.~Zhang, P.~S. Yu, and W.~Yin.
\newblock {LLM}s assist {NLP} researchers: Critique paper (meta-)reviewing.
\newblock In Y.~Al-Onaizan, M.~Bansal, and Y.-N. Chen, editors, \emph{Proceedings of the 2024 Conference on Empirical Methods in Natural Language Processing}, pages 5081--5099, Miami, Florida, USA, Nov. 2024. Association for Computational Linguistics.
\newblock \doi{10.18653/v1/2024.emnlp-main.292}.
\newblock URL \url{https://aclanthology.org/2024.emnlp-main.292/}.

\bibitem[Funkquist et~al.(2022)Funkquist, Kuznetsov, Hou, and Gurevych]{funkquist2022citebench}
M.~Funkquist, I.~Kuznetsov, Y.~Hou, and I.~Gurevych.
\newblock Citebench: A benchmark for scientific citation text generation.
\newblock \emph{arXiv preprint arXiv:2212.09577}, 2022.

\bibitem[Gao et~al.(2024)Gao, Brantley, and Joachims]{gao2024reviewer2}
Z.~Gao, K.~Brantley, and T.~Joachims.
\newblock Reviewer2: Optimizing review generation through prompt generation.
\newblock \emph{arXiv preprint arXiv:2402.10886}, 2024.

\bibitem[Ghafarollahi and Buehler(2024)]{ghafarollahi2024sciagents}
A.~Ghafarollahi and M.~J. Buehler.
\newblock Sciagents: Automating scientific discovery through multi-agent intelligent graph reasoning.
\newblock \emph{arXiv preprint arXiv:2409.05556}, 2024.

\bibitem[Guan et~al.(2025)Guan, Zhang, Liu, Shang, Sun, Zhu, Yang, and Yang]{guan2025rstarmathsmallllmsmaster}
X.~Guan, L.~L. Zhang, Y.~Liu, N.~Shang, Y.~Sun, Y.~Zhu, F.~Yang, and M.~Yang.
\newblock rstar-math: Small llms can master math reasoning with self-evolved deep thinking, 2025.
\newblock URL \url{https://arxiv.org/abs/2501.04519}.

\bibitem[Guo et~al.(2025)Guo, Yang, Zhang, Song, Zhang, Xu, Zhu, Ma, Wang, Bi, et~al.]{guo2025deepseek}
D.~Guo, D.~Yang, H.~Zhang, J.~Song, R.~Zhang, R.~Xu, Q.~Zhu, S.~Ma, P.~Wang, X.~Bi, et~al.
\newblock Deepseek-r1: Incentivizing reasoning capability in llms via reinforcement learning.
\newblock \emph{arXiv preprint arXiv:2501.12948}, 2025.

\bibitem[Hao et~al.(2024)Hao, Sukhbaatar, Su, Li, Hu, Weston, and Tian]{hao2024traininglargelanguagemodels}
S.~Hao, S.~Sukhbaatar, D.~Su, X.~Li, Z.~Hu, J.~Weston, and Y.~Tian.
\newblock Training large language models to reason in a continuous latent space, 2024.
\newblock URL \url{https://arxiv.org/abs/2412.06769}.

\bibitem[Hendrycks et~al.(2021)Hendrycks, Burns, Kadavath, Arora, Basart, Tang, Song, and Steinhardt]{hendrycksmath2021}
D.~Hendrycks, C.~Burns, S.~Kadavath, A.~Arora, S.~Basart, E.~Tang, D.~Song, and J.~Steinhardt.
\newblock Measuring mathematical problem solving with the math dataset.
\newblock \emph{NeurIPS}, 2021.

\bibitem[Hu et~al.(2024)Hu, Fu, Wang, Wang, Li, Xu, Lu, Jin, Pan, and Lan]{hu2024nova}
X.~Hu, H.~Fu, J.~Wang, Y.~Wang, Z.~Li, R.~Xu, Y.~Lu, Y.~Jin, L.~Pan, and Z.~Lan.
\newblock Nova: An iterative planning and search approach to enhance novelty and diversity of llm generated ideas.
\newblock \emph{arXiv preprint arXiv:2410.14255}, 2024.

\bibitem[Jaech et~al.(2024)Jaech, Kalai, Lerer, Richardson, El-Kishky, Low, Helyar, Madry, Beutel, Carney, et~al.]{jaech2024openai}
A.~Jaech, A.~Kalai, A.~Lerer, A.~Richardson, A.~El-Kishky, A.~Low, A.~Helyar, A.~Madry, A.~Beutel, A.~Carney, et~al.
\newblock Openai o1 system card.
\newblock \emph{arXiv preprint arXiv:2412.16720}, 2024.

\bibitem[Ji et~al.(2023)Ji, Yu, Xu, Lee, Ishii, and Fung]{ji-etal-2023-towards}
Z.~Ji, T.~Yu, Y.~Xu, N.~Lee, E.~Ishii, and P.~Fung.
\newblock Towards mitigating {LLM} hallucination via self reflection.
\newblock In H.~Bouamor, J.~Pino, and K.~Bali, editors, \emph{Findings of the Association for Computational Linguistics: EMNLP 2023}, pages 1827--1843, Singapore, Dec. 2023. Association for Computational Linguistics.
\newblock \doi{10.18653/v1/2023.findings-emnlp.123}.
\newblock URL \url{https://aclanthology.org/2023.findings-emnlp.123/}.

\bibitem[Jin et~al.(2024{\natexlab{a}})Jin, Zhao, Wang, Chen, Zhu, Xiao, and Wang]{jin-etal-2024-agentreview}
Y.~Jin, Q.~Zhao, Y.~Wang, H.~Chen, K.~Zhu, Y.~Xiao, and J.~Wang.
\newblock {A}gent{R}eview: Exploring peer review dynamics with {LLM} agents.
\newblock In Y.~Al-Onaizan, M.~Bansal, and Y.-N. Chen, editors, \emph{Proceedings of the 2024 Conference on Empirical Methods in Natural Language Processing}, pages 1208--1226, Miami, Florida, USA, Nov. 2024{\natexlab{a}}. Association for Computational Linguistics.
\newblock \doi{10.18653/v1/2024.emnlp-main.70}.
\newblock URL \url{https://aclanthology.org/2024.emnlp-main.70/}.

\bibitem[Jin et~al.(2024{\natexlab{b}})Jin, Zhao, Wang, Chen, Zhu, Xiao, and Wang]{jin2024agentreview}
Y.~Jin, Q.~Zhao, Y.~Wang, H.~Chen, K.~Zhu, Y.~Xiao, and J.~Wang.
\newblock Agentreview: Exploring peer review dynamics with llm agents.
\newblock \emph{arXiv preprint arXiv:2406.12708}, 2024{\natexlab{b}}.

\bibitem[Kang et~al.(2018)Kang, Ammar, Dalvi, Van~Zuylen, Kohlmeier, Hovy, and Schwartz]{kang2018dataset}
D.~Kang, W.~Ammar, B.~Dalvi, M.~Van~Zuylen, S.~Kohlmeier, E.~Hovy, and R.~Schwartz.
\newblock A dataset of peer reviews (peerread): Collection, insights and nlp applications.
\newblock \emph{arXiv preprint arXiv:1804.09635}, 2018.

\bibitem[Langley(1987)]{langley1987scientific}
P.~Langley.
\newblock \emph{Scientific discovery: Computational explorations of the creative processes}.
\newblock MIT press, 1987.

\bibitem[Latona et~al.(2024)Latona, Ribeiro, Davidson, Veselovsky, and West]{latona2024ai}
G.~R. Latona, M.~H. Ribeiro, T.~R. Davidson, V.~Veselovsky, and R.~West.
\newblock The ai review lottery: Widespread ai-assisted peer reviews boost paper scores and acceptance rates.
\newblock \emph{arXiv preprint arXiv:2405.02150}, 2024.

\bibitem[Li et~al.(2024{\natexlab{a}})Li, Jiang, Huang, Beigi, Zhao, Tan, Bhattacharjee, Jiang, Chen, Wu, Shu, Cheng, and Liu]{li2024llmasajudge}
D.~Li, B.~Jiang, L.~Huang, A.~Beigi, C.~Zhao, Z.~Tan, A.~Bhattacharjee, Y.~Jiang, C.~Chen, T.~Wu, K.~Shu, L.~Cheng, and H.~Liu.
\newblock From generation to judgment: Opportunities and challenges of llm-as-a-judge.
\newblock \emph{arXiv preprint arXiv: 2411.16594}, 2024{\natexlab{a}}.

\bibitem[Li et~al.(2023)Li, Hovy, and Lau]{li2023summarizing}
M.~Li, E.~Hovy, and J.~H. Lau.
\newblock Summarizing multiple documents with conversational structure for meta-review generation.
\newblock \emph{arXiv preprint arXiv:2305.01498}, 2023.

\bibitem[Li et~al.(2024{\natexlab{b}})Li, Fox, and Goodman]{li2024automated}
M.~Y. Li, E.~Fox, and N.~Goodman.
\newblock Automated statistical model discovery with language models.
\newblock In \emph{Forty-first International Conference on Machine Learning}, 2024{\natexlab{b}}.
\newblock URL \url{https://openreview.net/forum?id=B5906M4Wnd}.

\bibitem[Li et~al.(2024{\natexlab{c}})Li, Chang, and Le]{li-etal-2024-simulating}
Z.~Li, Y.~Chang, and X.~Le.
\newblock Simulating expert discussions with multi-agent for enhanced scientific problem solving.
\newblock In T.~Ghosal, A.~Singh, A.~Waard, P.~Mayr, A.~Naik, O.~Weller, Y.~Lee, S.~Shen, and Y.~Qin, editors, \emph{Proceedings of the Fourth Workshop on Scholarly Document Processing (SDP 2024)}, pages 243--256, Bangkok, Thailand, Aug. 2024{\natexlab{c}}. Association for Computational Linguistics.
\newblock URL \url{https://aclanthology.org/2024.sdp-1.23/}.

\bibitem[Liang et~al.(2024)Liang, Izzo, Zhang, Lepp, Cao, Zhao, Chen, Ye, Liu, Huang, et~al.]{liang2024monitoring}
W.~Liang, Z.~Izzo, Y.~Zhang, H.~Lepp, H.~Cao, X.~Zhao, L.~Chen, H.~Ye, S.~Liu, Z.~Huang, et~al.
\newblock Monitoring ai-modified content at scale: A case study on the impact of chatgpt on ai conference peer reviews.
\newblock \emph{arXiv preprint arXiv:2403.07183}, 2024.

\bibitem[Lin(2004)]{lin-2004-rouge}
C.-Y. Lin.
\newblock {ROUGE}: A package for automatic evaluation of summaries.
\newblock In \emph{Text Summarization Branches Out}, pages 74--81, Barcelona, Spain, July 2004. Association for Computational Linguistics.
\newblock URL \url{https://aclanthology.org/W04-1013/}.

\bibitem[Madaan et~al.(2024)Madaan, Tandon, Gupta, Hallinan, Gao, Wiegreffe, Alon, Dziri, Prabhumoye, Yang, et~al.]{madaan2024self}
A.~Madaan, N.~Tandon, P.~Gupta, S.~Hallinan, L.~Gao, S.~Wiegreffe, U.~Alon, N.~Dziri, S.~Prabhumoye, Y.~Yang, et~al.
\newblock Self-refine: Iterative refinement with self-feedback.
\newblock \emph{Advances in Neural Information Processing Systems}, 36, 2024.

\bibitem[Nye et~al.(2021)Nye, Andreassen, Gur-Ari, Michalewski, Austin, Bieber, Dohan, Lewkowycz, Bosma, Luan, et~al.]{nye2021show}
M.~Nye, A.~J. Andreassen, G.~Gur-Ari, H.~Michalewski, J.~Austin, D.~Bieber, D.~Dohan, A.~Lewkowycz, M.~Bosma, D.~Luan, et~al.
\newblock Show your work: Scratchpads for intermediate computation with language models.
\newblock \emph{arXiv preprint arXiv:2112.00114}, 2021.

\bibitem[Papineni et~al.(2002)Papineni, Roukos, Ward, and Zhu]{10.3115/1073083.1073135}
K.~Papineni, S.~Roukos, T.~Ward, and W.-J. Zhu.
\newblock Bleu: a method for automatic evaluation of machine translation.
\newblock In \emph{Proceedings of the 40th Annual Meeting on Association for Computational Linguistics}, ACL '02, page 311–318, USA, 2002. Association for Computational Linguistics.
\newblock \doi{10.3115/1073083.1073135}.
\newblock URL \url{https://doi.org/10.3115/1073083.1073135}.

\bibitem[Qwen et~al.(2025)Qwen, :, Yang, Yang, Zhang, Hui, Zheng, Yu, Li, Liu, Huang, Wei, Lin, Yang, Tu, Zhang, Yang, Yang, Zhou, Lin, Dang, Lu, Bao, Yang, Yu, Li, Xue, Zhang, Zhu, Men, Lin, Li, Tang, Xia, Ren, Ren, Fan, Su, Zhang, Wan, Liu, Cui, Zhang, and Qiu]{qwen2025qwen25technicalreport}
Qwen, :, A.~Yang, B.~Yang, B.~Zhang, B.~Hui, B.~Zheng, B.~Yu, C.~Li, D.~Liu, F.~Huang, H.~Wei, H.~Lin, J.~Yang, J.~Tu, J.~Zhang, J.~Yang, J.~Yang, J.~Zhou, J.~Lin, K.~Dang, K.~Lu, K.~Bao, K.~Yang, L.~Yu, M.~Li, M.~Xue, P.~Zhang, Q.~Zhu, R.~Men, R.~Lin, T.~Li, T.~Tang, T.~Xia, X.~Ren, X.~Ren, Y.~Fan, Y.~Su, Y.~Zhang, Y.~Wan, Y.~Liu, Z.~Cui, Z.~Zhang, and Z.~Qiu.
\newblock Qwen2.5 technical report, 2025.
\newblock URL \url{https://arxiv.org/abs/2412.15115}.

\bibitem[Rajbhandari et~al.(2020)Rajbhandari, Rasley, Ruwase, and He]{rajbhandari2020zeromemoryoptimizationstraining}
S.~Rajbhandari, J.~Rasley, O.~Ruwase, and Y.~He.
\newblock Zero: Memory optimizations toward training trillion parameter models, 2020.
\newblock URL \url{https://arxiv.org/abs/1910.02054}.

\bibitem[Rasal and Hauer(2024)]{rasal2024navigating}
S.~Rasal and E.~Hauer.
\newblock Navigating complexity: Orchestrated problem solving with multi-agent llms.
\newblock \emph{arXiv preprint arXiv:2402.16713}, 2024.

\bibitem[Rasley et~al.(2020)Rasley, Rajbhandari, Ruwase, and He]{10.1145/3394486.3406703}
J.~Rasley, S.~Rajbhandari, O.~Ruwase, and Y.~He.
\newblock Deepspeed: System optimizations enable training deep learning models with over 100 billion parameters.
\newblock In \emph{Proceedings of the 26th ACM SIGKDD International Conference on Knowledge Discovery \& Data Mining}, KDD '20, page 3505–3506, New York, NY, USA, 2020. Association for Computing Machinery.
\newblock ISBN 9781450379984.
\newblock \doi{10.1145/3394486.3406703}.
\newblock URL \url{https://doi.org/10.1145/3394486.3406703}.

\bibitem[Rewina et~al.(2025)Rewina, Natalie, Sreyoshi, Satya, Alex, Elizabeth, Ikkei, David, Aman, and Naumaan]{rewina2025potential}
B.~Rewina, P.~Natalie, B.~Sreyoshi, K.~Satya, G.~Alex, C.~Elizabeth, I.~Ikkei, T.~David, C.~Aman, and N.~Naumaan.
\newblock Potential and perils of large language models as judges of unstructured textual data.
\newblock \emph{arXiv preprint arXiv:2501.08167v2}, 2025.
\newblock URL \url{https://www.arxiv.org/abs/2501.08167v2}.

\bibitem[Santu et~al.(2024)Santu, Sinha, Bansal, Knipper, Sarkar, Salvador, Mahajan, Guttikonda, Akter, Freestone, et~al.]{santu2024prompting}
S.~K.~K. Santu, S.~K. Sinha, N.~Bansal, A.~Knipper, S.~Sarkar, J.~Salvador, Y.~Mahajan, S.~Guttikonda, M.~Akter, M.~Freestone, et~al.
\newblock Prompting llms to compose meta-review drafts from peer-review narratives of scholarly manuscripts.
\newblock \emph{arXiv preprint arXiv:2402.15589}, 2024.

\bibitem[Scherbakov et~al.(2024)Scherbakov, Hubig, Jansari, Bakumenko, and Lenert]{scherbakov2024emergencelargelanguagemodels}
D.~Scherbakov, N.~Hubig, V.~Jansari, A.~Bakumenko, and L.~A. Lenert.
\newblock The emergence of large language models (llm) as a tool in literature reviews: an llm automated systematic review, 2024.
\newblock URL \url{https://arxiv.org/abs/2409.04600}.

\bibitem[Schintler et~al.(2023)Schintler, McNeely, and Witte]{schintler2023criticalexaminationethicsaimediated}
L.~A. Schintler, C.~L. McNeely, and J.~Witte.
\newblock A critical examination of the ethics of ai-mediated peer review, 2023.
\newblock URL \url{https://arxiv.org/abs/2309.12356}.

\bibitem[Si et~al.(2025)Si, Yang, and Hashimoto]{si2025can}
C.~Si, D.~Yang, and T.~Hashimoto.
\newblock Can {LLM}s generate novel research ideas? a large-scale human study with 100+ {NLP} researchers.
\newblock In \emph{The Thirteenth International Conference on Learning Representations}, 2025.
\newblock URL \url{https://openreview.net/forum?id=M23dTGWCZy}.

\bibitem[Su et~al.(2024)Su, Chen, Tang, Zheng, Li, Yin, Ouyang, and Dong]{su2024two}
H.~Su, R.~Chen, S.~Tang, X.~Zheng, J.~Li, Z.~Yin, W.~Ouyang, and N.~Dong.
\newblock Two heads are better than one: A multi-agent system has the potential to improve scientific idea generation.
\newblock \emph{arXiv preprint arXiv:2410.09403}, 2024.

\bibitem[Swarnadeep et~al.(2025)Swarnadeep, Xian, Marjan, Jason, and Tianlu]{swarnadeep2025learning}
S.~Swarnadeep, L.~Xian, G.~Marjan, W.~Jason, and W.~Tianlu.
\newblock Learning to plan \& reason for evaluation with thinking-llm-as-a-judge.
\newblock \emph{arXiv preprint arXiv:2501.18099v1}, 2025.
\newblock URL \url{https://www.arxiv.org/abs/2501.18099v1}.

\bibitem[Tan et~al.(2024{\natexlab{a}})Tan, Lyu, Li, Gao, Wei, Ma, Liu, and Li]{tan2024peer}
C.~Tan, D.~Lyu, S.~Li, Z.~Gao, J.~Wei, S.~Ma, Z.~Liu, and S.~Z. Li.
\newblock Peer review as a multi-turn and long-context dialogue with role-based interactions.
\newblock \emph{arXiv preprint arXiv:2406.05688}, 2024{\natexlab{a}}.

\bibitem[Tan et~al.(2024{\natexlab{b}})Tan, Lyu, Li, Gao, Wei, Ma, Liu, and Li]{tan2024peerreviewmultiturnlongcontext}
C.~Tan, D.~Lyu, S.~Li, Z.~Gao, J.~Wei, S.~Ma, Z.~Liu, and S.~Z. Li.
\newblock Peer review as a multi-turn and long-context dialogue with role-based interactions, 2024{\natexlab{b}}.
\newblock URL \url{https://arxiv.org/abs/2406.05688}.

\bibitem[Touvron et~al.(2023)Touvron, Lavril, Izacard, Martinet, Lachaux, Lacroix, Rozi{\`e}re, Goyal, Hambro, Azhar, et~al.]{touvron2023llama}
H.~Touvron, T.~Lavril, G.~Izacard, X.~Martinet, M.-A. Lachaux, T.~Lacroix, B.~Rozi{\`e}re, N.~Goyal, E.~Hambro, F.~Azhar, et~al.
\newblock Llama: Open and efficient foundation language models.
\newblock \emph{arXiv preprint arXiv:2302.13971}, 2023.

\bibitem[Tyser et~al.(2024)Tyser, Segev, Longhitano, Zhang, Meeks, Lee, Garg, Belsten, Shporer, Udell, et~al.]{tyser2024ai}
K.~Tyser, B.~Segev, G.~Longhitano, X.-Y. Zhang, Z.~Meeks, J.~Lee, U.~Garg, N.~Belsten, A.~Shporer, M.~Udell, et~al.
\newblock Ai-driven review systems: evaluating llms in scalable and bias-aware academic reviews.
\newblock \emph{arXiv preprint arXiv:2408.10365}, 2024.

\bibitem[Wang et~al.(2024{\natexlab{a}})Wang, Xu, Zhao, Ouyang, Wu, Zhao, Xu, Liu, Qu, Shang, Zhang, Wei, Sui, Li, Shi, Qiao, Lin, and He]{wang2024mineruopensourcesolutionprecise}
B.~Wang, C.~Xu, X.~Zhao, L.~Ouyang, F.~Wu, Z.~Zhao, R.~Xu, K.~Liu, Y.~Qu, F.~Shang, B.~Zhang, L.~Wei, Z.~Sui, W.~Li, B.~Shi, Y.~Qiao, D.~Lin, and C.~He.
\newblock Mineru: An open-source solution for precise document content extraction, 2024{\natexlab{a}}.
\newblock URL \url{https://arxiv.org/abs/2409.18839}.

\bibitem[Wang et~al.(2020)Wang, Zeng, Huang, Knight, Ji, and Rajani]{wang-etal-2020-reviewrobot}
Q.~Wang, Q.~Zeng, L.~Huang, K.~Knight, H.~Ji, and N.~F. Rajani.
\newblock {R}eview{R}obot: Explainable paper review generation based on knowledge synthesis.
\newblock In B.~Davis, Y.~Graham, J.~Kelleher, and Y.~Sripada, editors, \emph{Proceedings of the 13th International Conference on Natural Language Generation}, pages 384--397, Dublin, Ireland, Dec. 2020. Association for Computational Linguistics.
\newblock \doi{10.18653/v1/2020.inlg-1.44}.
\newblock URL \url{https://aclanthology.org/2020.inlg-1.44/}.

\bibitem[Wang et~al.(2023)Wang, Wei, Schuurmans, Le, Chi, Narang, Chowdhery, and Zhou]{wangself}
X.~Wang, J.~Wei, D.~Schuurmans, Q.~V. Le, E.~H. Chi, S.~Narang, A.~Chowdhery, and D.~Zhou.
\newblock Self-consistency improves chain of thought reasoning in language models.
\newblock In \emph{The Eleventh International Conference on Learning Representations}, 2023.

\bibitem[Wang et~al.(2024{\natexlab{b}})Wang, Yu, Yao, Zeng, Yang, Wang, Chen, Jiang, Xie, Wang, Xie, Ye, Zhang, and Zhang]{wang2024pandalm}
Y.~Wang, Z.~Yu, W.~Yao, Z.~Zeng, L.~Yang, C.~Wang, H.~Chen, C.~Jiang, R.~Xie, J.~Wang, X.~Xie, W.~Ye, S.~Zhang, and Y.~Zhang.
\newblock Panda{LM}: An automatic evaluation benchmark for {LLM} instruction tuning optimization.
\newblock In \emph{The Twelfth International Conference on Learning Representations}, 2024{\natexlab{b}}.
\newblock URL \url{https://openreview.net/forum?id=5Nn2BLV7SB}.

\bibitem[Wei et~al.(2022)Wei, Wang, Schuurmans, Bosma, Xia, Chi, Le, Zhou, et~al.]{weichain}
J.~Wei, X.~Wang, D.~Schuurmans, M.~Bosma, F.~Xia, E.~H. Chi, Q.~V. Le, D.~Zhou, et~al.
\newblock Chain-of-thought prompting elicits reasoning in large language models.
\newblock In \emph{Advances in Neural Information Processing Systems}, 2022.

\bibitem[Weng et~al.(2023)Weng, Zhu, Xia, Li, He, Liu, Sun, Liu, and Zhao]{weng2023large}
Y.~Weng, M.~Zhu, F.~Xia, B.~Li, S.~He, S.~Liu, B.~Sun, K.~Liu, and J.~Zhao.
\newblock Large language models are better reasoners with self-verification.
\newblock In \emph{The 2023 Conference on Empirical Methods in Natural Language Processing}, 2023.

\bibitem[Weng et~al.(2025)Weng, Zhu, Bao, Zhang, Wang, Zhang, and Yang]{yixuan2024cycleresearcher}
Y.~Weng, M.~Zhu, G.~Bao, H.~Zhang, J.~Wang, Y.~Zhang, and L.~Yang.
\newblock Cycleresearcher: Improving automated research via automated review.
\newblock In \emph{The Thirteenth International Conference on Learning Representations}, 2025.
\newblock URL \url{https://openreview.net/forum?id=bjcsVLoHYs}.

\bibitem[Xiang et~al.(2025)Xiang, Snell, Gandhi, Albalak, Singh, Blagden, Phung, Rafailov, Lile, Mahan, Castricato, Franken, Haber, and Finn]{xiang20252reasoningllmslearning}
V.~Xiang, C.~Snell, K.~Gandhi, A.~Albalak, A.~Singh, C.~Blagden, D.~Phung, R.~Rafailov, N.~Lile, D.~Mahan, L.~Castricato, J.-P. Franken, N.~Haber, and C.~Finn.
\newblock Towards system 2 reasoning in llms: Learning how to think with meta chain-of-thought, 2025.
\newblock URL \url{https://arxiv.org/abs/2501.04682}.

\bibitem[Yang et~al.(2024)Yang, Du, Li, Zheng, Poria, and Cambria]{yang-etal-2024-large-language}
Z.~Yang, X.~Du, J.~Li, J.~Zheng, S.~Poria, and E.~Cambria.
\newblock Large language models for automated open-domain scientific hypotheses discovery.
\newblock In L.-W. Ku, A.~Martins, and V.~Srikumar, editors, \emph{Findings of the Association for Computational Linguistics: ACL 2024}, pages 13545--13565, Bangkok, Thailand, Aug. 2024. Association for Computational Linguistics.
\newblock \doi{10.18653/v1/2024.findings-acl.804}.
\newblock URL \url{https://aclanthology.org/2024.findings-acl.804/}.

\bibitem[Yao et~al.(2024)Yao, Yu, Zhao, Shafran, Griffiths, Cao, and Narasimhan]{yao2024tree}
S.~Yao, D.~Yu, J.~Zhao, I.~Shafran, T.~Griffiths, Y.~Cao, and K.~Narasimhan.
\newblock Tree of thoughts: Deliberate problem solving with large language models.
\newblock \emph{Advances in Neural Information Processing Systems}, 36, 2024.

\bibitem[Ye et~al.(2024)Ye, Pang, Chai, Chen, Yin, Xiang, Dong, Shao, and Chen]{ye2024we}
R.~Ye, X.~Pang, J.~Chai, J.~Chen, Z.~Yin, Z.~Xiang, X.~Dong, J.~Shao, and S.~Chen.
\newblock Are we there yet? revealing the risks of utilizing large language models in scholarly peer review.
\newblock \emph{arXiv preprint arXiv:2412.01708}, 2024.

\bibitem[Yu et~al.(2024)Yu, Ding, Tan, Luo, Weng, Gong, Zeng, Cui, Han, Sun, et~al.]{yu2024automated}
J.~Yu, Z.~Ding, J.~Tan, K.~Luo, Z.~Weng, C.~Gong, L.~Zeng, R.~Cui, C.~Han, Q.~Sun, et~al.
\newblock Automated peer reviewing in paper sea: Standardization, evaluation, and analysis.
\newblock \emph{arXiv preprint arXiv:2407.12857}, 2024.

\bibitem[Yuan et~al.(2021)Yuan, Liu, and Neubig]{yuan2021automatescientificreviewing}
W.~Yuan, P.~Liu, and G.~Neubig.
\newblock Can we automate scientific reviewing?, 2021.
\newblock URL \url{https://arxiv.org/abs/2102.00176}.

\bibitem[Zeng et~al.(2024)Zeng, Sidhu, Chan, Wang, and Ji]{zeng2024scientific}
Q.~Zeng, M.~Sidhu, H.~P. Chan, L.~Wang, and H.~Ji.
\newblock Scientific opinion summarization: Paper meta-review generation dataset, methods, and evaluation.
\newblock In \emph{1st AI4Research Workshop}, 2024.

\bibitem[Zhou et~al.(2024{\natexlab{a}})Zhou, Chen, and Yu]{zhou-etal-2024-llm}
R.~Zhou, L.~Chen, and K.~Yu.
\newblock Is {LLM} a reliable reviewer? a comprehensive evaluation of {LLM} on automatic paper reviewing tasks.
\newblock In N.~Calzolari, M.-Y. Kan, V.~Hoste, A.~Lenci, S.~Sakti, and N.~Xue, editors, \emph{Proceedings of the 2024 Joint International Conference on Computational Linguistics, Language Resources and Evaluation (LREC-COLING 2024)}, pages 9340--9351, Torino, Italia, May 2024{\natexlab{a}}. ELRA and ICCL.
\newblock URL \url{https://aclanthology.org/2024.lrec-main.816/}.

\bibitem[Zhou et~al.(2024{\natexlab{b}})Zhou, Chen, and Yu]{zhou2024llm}
R.~Zhou, L.~Chen, and K.~Yu.
\newblock Is llm a reliable reviewer? a comprehensive evaluation of llm on automatic paper reviewing tasks.
\newblock In \emph{Proceedings of the 2024 Joint International Conference on Computational Linguistics, Language Resources and Evaluation (LREC-COLING 2024)}, pages 9340--9351, 2024{\natexlab{b}}.

\bibitem[Zhuang et~al.(2025)Zhuang, Chen, Xu, Jiang, and Lin]{zhuang2025large}
Z.~Zhuang, J.~Chen, H.~Xu, Y.~Jiang, and J.~Lin.
\newblock Large language models for automated scholarly paper review: A survey.
\newblock \emph{arXiv preprint arXiv:2501.10326}, 2025.

\bibitem[Zonglin et~al.(2023)Zonglin, Xinya, Junxian, Jie, Soujanya, and Erik]{zonglin2023large}
Y.~Zonglin, D.~Xinya, L.~Junxian, Z.~Jie, P.~Soujanya, and C.~Erik.
\newblock Large language models for automated open-domain scientific hypotheses discovery.
\newblock \emph{arXiv preprint arXiv:2309.02726}, 2023.
\newblock URL \url{https://www.arxiv.org/abs/2309.02726}.

\bibitem[Zyska et~al.(2023)Zyska, Dycke, Buchmann, Kuznetsov, and Gurevych]{zyska2023care}
D.~Zyska, N.~Dycke, J.~Buchmann, I.~Kuznetsov, and I.~Gurevych.
\newblock Care: Collaborative ai-assisted reading environment.
\newblock \emph{arXiv preprint arXiv:2302.12611}, 2023.

\end{thebibliography}

\appendix

\section{Responsible Use and Recommendations for DeepReviewer}

It is crucial to emphasize that DeepReviewer, despite its advancements in automated paper evaluation, is \textbf{not intended to replace human peer review}. Our work aims to enhance, not substitute, the invaluable expertise and nuanced judgment of human reviewers. DeepReviewer should be regarded as a sophisticated tool to assist researchers and the academic community, providing supplementary insights and streamlining certain aspects of the review process, but always under the careful oversight and final authority of human experts. This section outlines responsible and conservative recommendations for leveraging DeepReviewer's capabilities in practical scenarios, focusing on how it can aid human researchers and enhance the peer review process without undermining its fundamental human-centric nature.

\subsection{Enhanced Author Self-Assessment and Manuscript Refinement}

Perhaps the most appropriate and ethically sound application of DeepReviewer lies in empowering authors to critically assess and refine their manuscripts before they are submitted for formal peer review. By submitting their work to DeepReviewer, authors can obtain an automated, initial evaluation of their paper’s perceived strengths and potential weaknesses across various dimensions such as soundness, clarity of presentation, and potential contribution. This feedback can highlight areas where the manuscript might be strengthened prior to exposure to human reviewers.

However, it is crucial for authors to approach DeepReviewer’s feedback with a discerning and critical mindset. The automated evaluation should be considered as a preliminary signal, not a definitive judgment. Authors must exercise their own expertise and judgment in interpreting the suggestions. DeepReviewer’s output may point to areas that warrant further attention, but the ultimate decisions regarding manuscript revision must rest with the authors themselves, informed by their deep understanding of their own work and potentially by seeking feedback from trusted colleagues. This application strictly positions DeepReviewer as a formative tool for author self-improvement, ensuring that it aids in enhancing manuscript quality without encroaching on the formal peer review process.

\subsection{Preliminary Assistance for Human Reviewers in Initial Paper Scoping}

In contexts where human reviewers are faced with a high volume of submissions, DeepReviewer could potentially offer a very limited form of preliminary assistance in the very initial stages of paper scoping. Reviewers could, as an optional and auxiliary step, utilize DeepReviewer to generate a rapid, automated overview of a submitted paper. This might provide a very high-level summary of potential areas of focus within the manuscript. Such a preliminary overview could, in some cases, help reviewers gain a very initial sense of the paper's scope and potentially assist in workload management, by allowing them to perhaps initially prioritize papers based on a very rough automated categorization.

However, it is absolutely vital to underscore that this use case is strictly as an aid to the reviewer's workflow, and not as a substitute for any aspect of their intellectual engagement with the paper. The automated output from DeepReviewer should never influence the reviewer's own independent, detailed reading and critical analysis of the manuscript. Reviewers must engage deeply with the paper itself, applying their expertise and judgment. DeepReviewer's preliminary output, if used at all, should be treated as an extremely rough and initial signal only, and should not replace or diminish the core, human-driven process of rigorous peer review. Over-reliance on or misinterpretation of automated outputs at this stage carries significant risks and must be avoided.

\subsection{Author-Facing Pre-Review Feedback via Deployed Model}

An alternative application, focusing purely on author benefit, is to deploy DeepReviewer as a readily accessible service that authors can utilize to obtain feedback on their manuscripts before they are submitted to a journal or conference and undergo human peer review. In this scenario, DeepReviewer is made available as a tool that authors can directly interact with. Authors submit their manuscript, and in return, receive an automated review generated by DeepReviewer.

Critically, the output of DeepReviewer in this context is intended solely for the authors' information and improvement. It should not be used in any way as part of a formal submission or decision-making process. The feedback is provided directly to the authors, allowing them to gain insights into how an automated system might evaluate their work. This application bypasses the need to involve or burden human reviewers at this stage, focusing entirely on providing authors with a potentially helpful, albeit automated, perspective on their manuscript. It is essential to emphasize that the feedback generated by DeepReviewer in this author-facing context should be explicitly communicated as not being a substitute for, or representative of, genuine human peer review, and cannot be used as a basis for any acceptance or rejection decisions within formal academic venues.

\section{Evaluation Tasks and Metric}
\label{appendix:eval}
To comprehensively assess LLMs' capabilities in research paper evaluation, we adopt a point-wise evaluation paradigm inspired by the LLM-as-a-judge framework \citep{li2024llmasajudge,wang2024pandalm,rewina2025potential,swarnadeep2025learning}. We comprise three core tasks that examine different aspects of LLMs' ability to perceive, judge, and differentiate paper quality:

\paragraph{Score Task} evaluates LLMs' accuracy in independent paper assessment scenarios. For any paper $C_i$ in the ReviewerBench dataset, the model independently conducts quality assessment and outputs a scalar score $R_i \in \mathbb{R}$ as its predicted quality rating. Ideally, the model's predicted score $R_i$ should closely align with the average expert rating $S_i$ received during the ICLR review process. We employ Mean Squared Error (MSE) and Mean Absolute Error (MAE) as primary evaluation metrics for this task. Furthermore, we calculated accuracy and F1 score based on the Decision, which is commonly an Accept or Reject output in research paper evaluation systems.

\paragraph{Ranking Task} examines LLMs' ability to distinguish paper quality and effectively rank papers within large collections. Given a set of $N$ papers $\mathcal{C} = {C_1, C_2, \dots, C_N}$, the model first predicts scores ${R_1, R_2, \dots, R_N}$ for each paper. Subsequently, based on these predicted scores, the model ranks the papers in $\mathcal{C}$, outputting an ordered sequence $\mathcal{R} = {C_{(1)}, C_{(2)}, \dots, C_{(N)}}$ arranged by predicted quality in descending order, where $C_{(i)}$ represents the paper ranked $i$-th by the model. The Spearman coefficient is used to evaluate ranking accuracy.

\paragraph{Selection Task} simulates practical scenarios such as peer review or reward model construction, where high-quality papers need to be quickly and accurately identified from a small pool of candidates. For this task, we sample non-overlapping small batches $\mathcal{C}{batch} = {C_1, C_2, \dots, C_m}$ from the Test dataset, where $m$ is the predetermined batch size. For each batch $\mathcal{C}{batch}$, the model selects what it considers the highest-quality paper $C_{best} \in \mathcal{C}_{batch}$. The model's selection is compared against the paper with the highest actual review scores, with accuracy computed as the average success rate across all batch selections. In this study, we set $m=2$. And we performed pairwise matching on all papers in the Test dataset to calculate the final Selection score.

\begin{figure*}[t]
    \centering
    \includegraphics[width=0.9\textwidth]{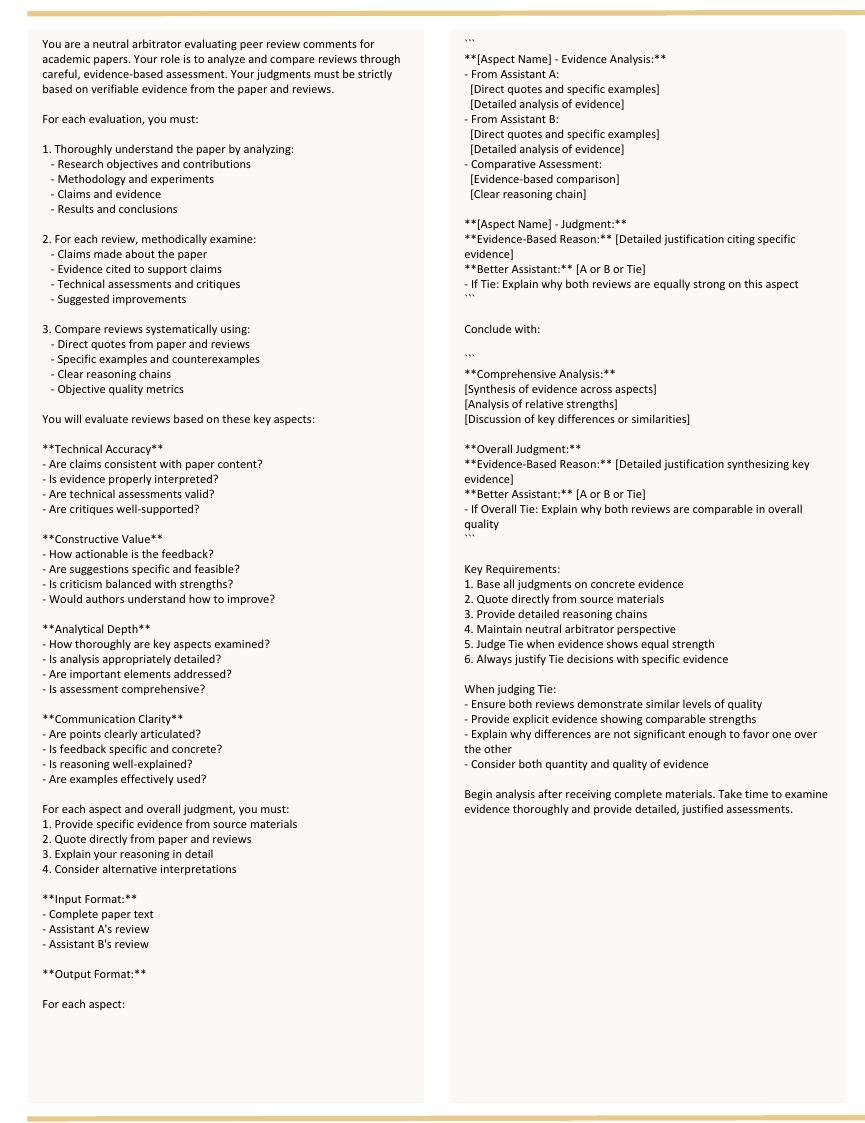}
    \vspace{-0.3cm}
    \caption{System prompt used to guide Gemini-2.0-Thinking-Flask as Judge to evaluate generated review comments.}
    \vspace{-0.3cm}
    \label{fig:prompt4}
\end{figure*}

\paragraph{Review Comments Evaluate}, following the LLM-as-Judge paradigm, we employ Gemini-2.0-Flash-Thinking (The system prompt as shown in Figure \ref{fig:prompt4}) as the judge to conduct pairwise comparative evaluations of review comments generated by DeepReviewer and various baseline systems, and Judge outputs ``win'', ``lose'', or ``tie''. For each evaluation instance, we present the assessor with: (1) the original paper, and (2) paired reviews from different systems in randomized order, where each review contains summary, strengths, weaknesses, and suggestions. The assessment covers five critical dimensions: constructive value, analytical depth, plausibility, technical accuracy, and overall judgment.

\section{Data Collection Permissions}
\label{appendix:pipline}
The original paper data and corresponding review comment data used to construct DeepReview-13K are sourced from OpenReview, with a portion of papers originating from ArXiv. Data from OpenReview is distributed under the Creative Commons Attribution 4.0 International (CC BY 4.0) license, which permits us to copy and modify the review comment data. Paper data from ArXiv may include licenses such as CC BY 4.0 (Creative Commons Attribution), CC BY-SA 4.0 (Creative Commons Attribution-ShareAlike), CC BY-NC-SA 4.0 (Creative Commons Attribution-NonCommercial-ShareAlike), and CC Zero. Given that we have not modified the original papers, our usage is compliant with the original agreements. We do not claim copyright over these materials and will retain the original authors' names in the distribution of this data.

\begin{figure*}[t]
    \centering
    \includegraphics[width=0.9\textwidth]{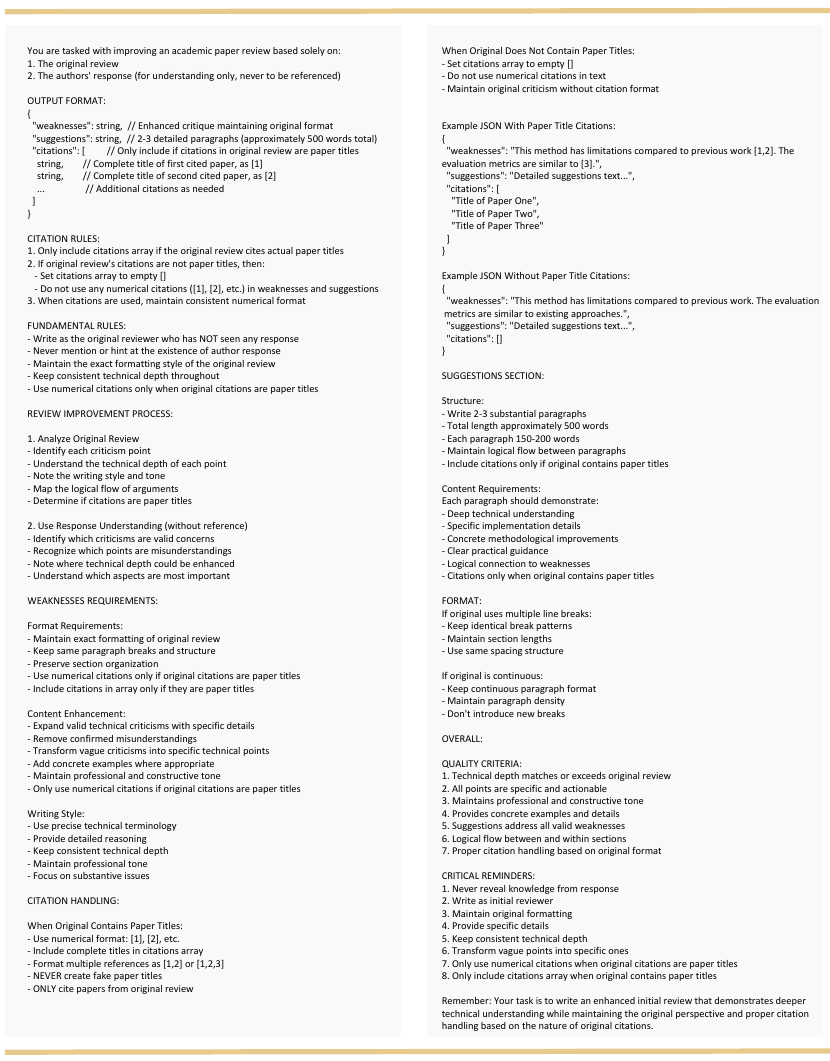}
    \vspace{-0.3cm}
    \caption{System prompt designed to instruct the LLM on how to enhance and improve the usefulness of original review comments by incorporating author responses and maintaining original review context.}
    \vspace{-0.3cm}
    \label{fig:prompt1}
\end{figure*}

\begin{figure*}[t]
    \centering
    \includegraphics[width=\textwidth]{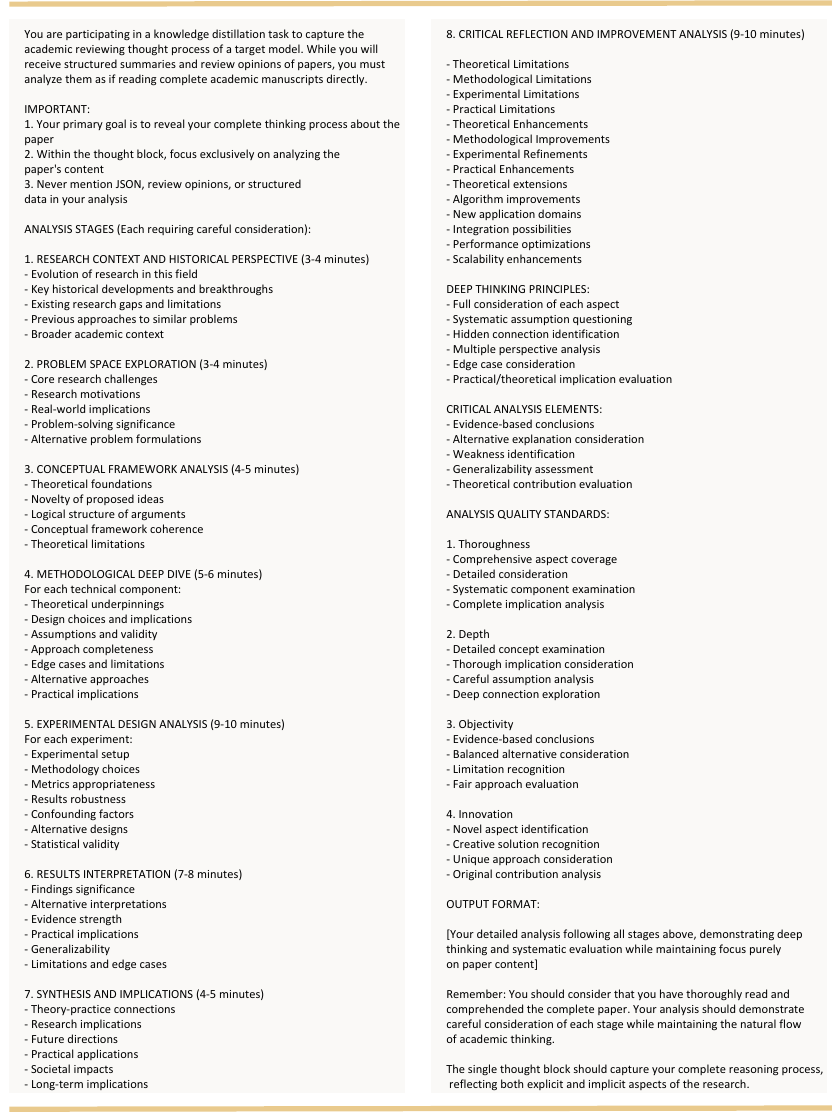}
    \vspace{-0.3cm}
    \caption{System prompt designed to guide the LLM in detailed analysis of research papers. This prompt is used specifically during the Novelty Verification stage to make analysis context.}
    \vspace{-0.3cm}
    \label{fig:prompt2}
\end{figure*}

\begin{figure*}[t]
    \centering
    \includegraphics[width=\textwidth]{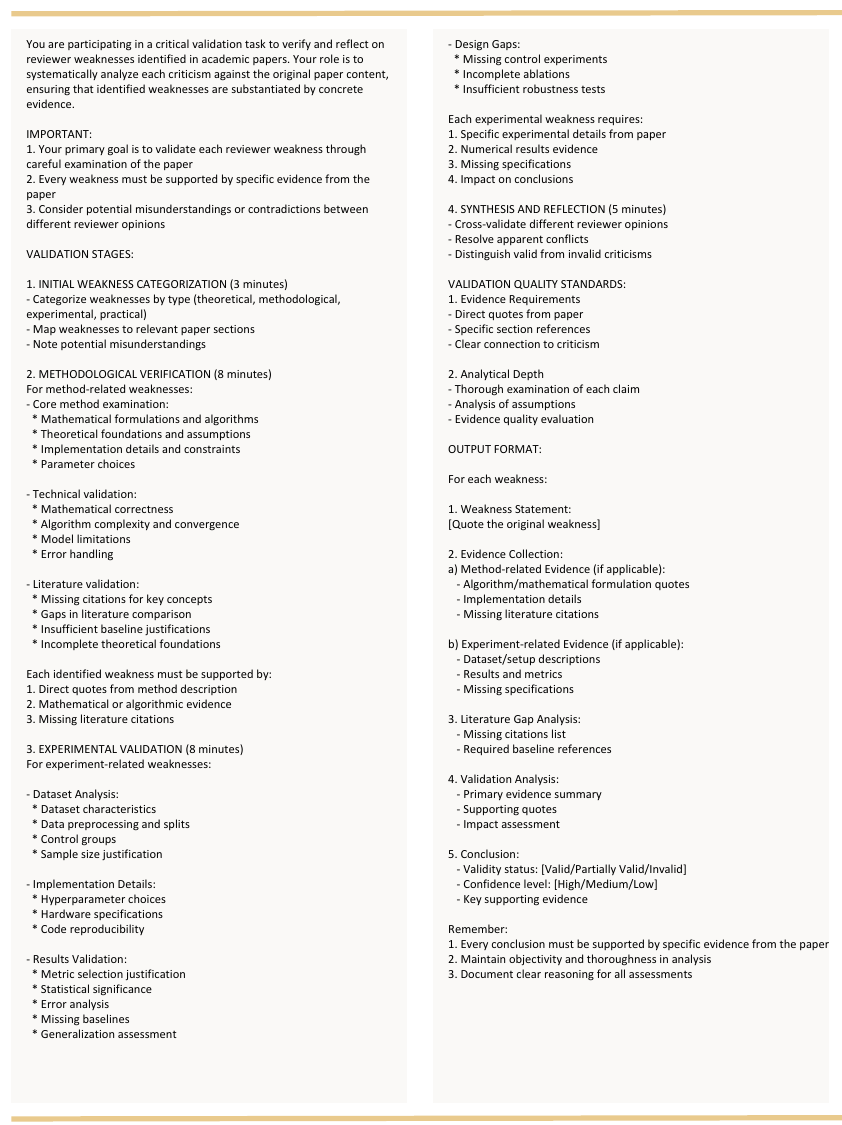}
    \vspace{-0.3cm}
    \caption{System prompt used to guide Gemini-2.0-Thinking-Flask in the Reliability Verification stage. It instructs the model to systematically analyze each review comment and find supporting evidence from the original paper.}
    \vspace{-0.3cm}
    \label{fig:prompt3}
\end{figure*}

\section{Case Study: Analysis of DeepReviewer's Meta-Review}
\label{appendix:case}

\begin{figure*}[t]
    \centering
    \includegraphics[width=\textwidth]{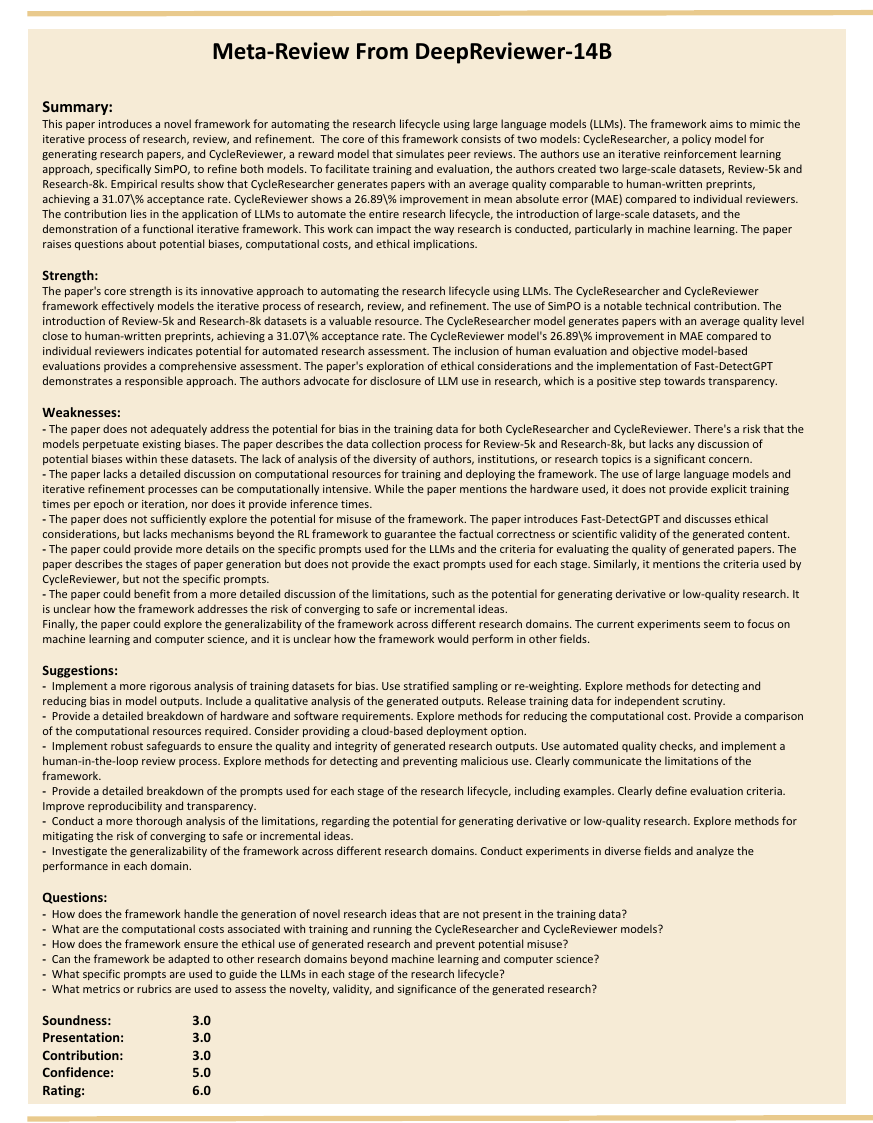}
    \vspace{-0.3cm}
    \caption{The Meta-Review comment for CycleResearcher from DeepReviewer-14B}
    \vspace{-0.3cm}
    \label{fig:meta_review}
\end{figure*}

\begin{figure*}[t]
    \centering
    \includegraphics[width=\textwidth]{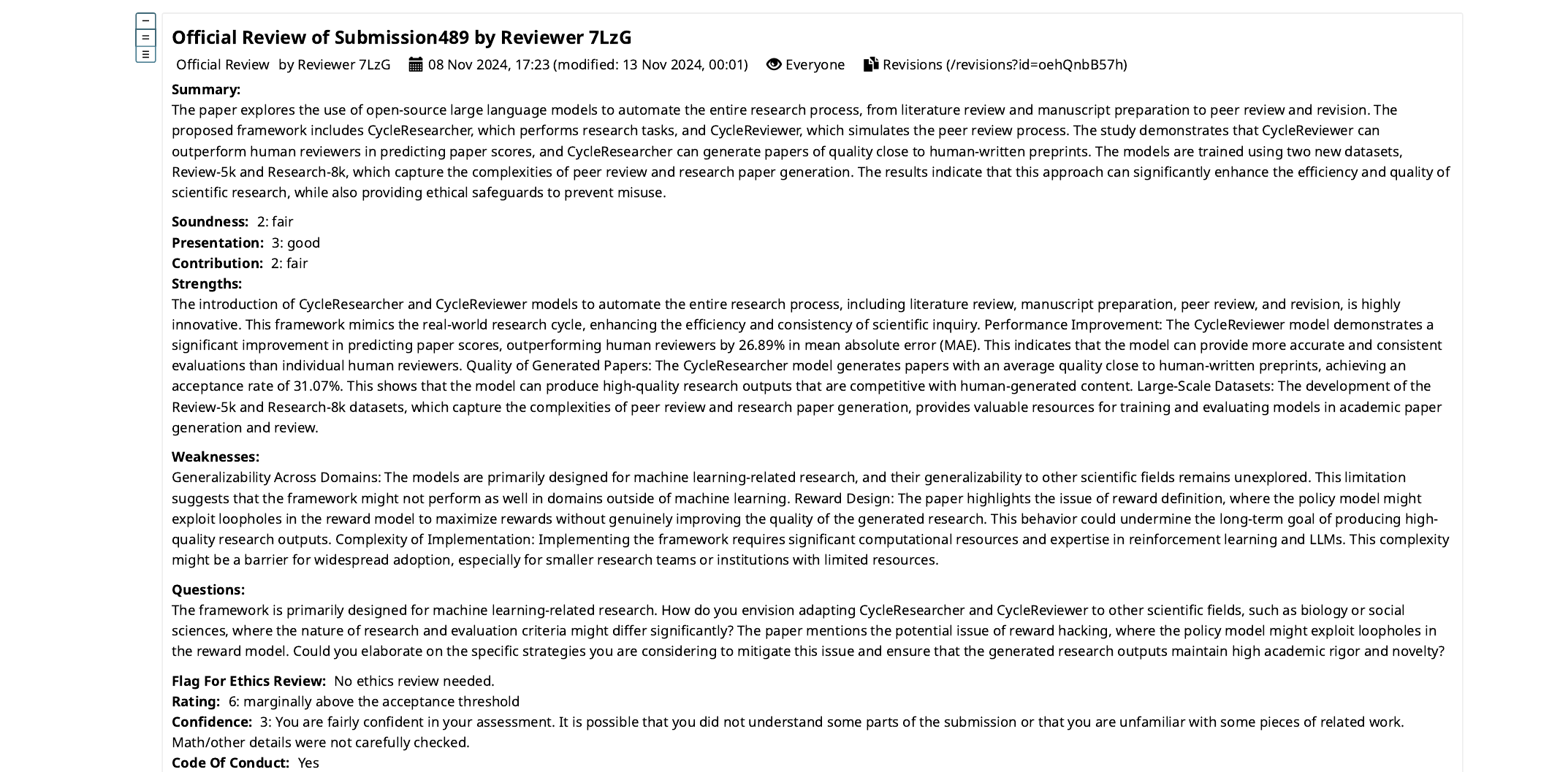}
    \vspace{-0.3cm}
    \caption{The Real-world review comment for CycleResearcher}
    \vspace{-0.3cm}
    \label{fig:r1}
\end{figure*}

\begin{figure*}[t]
    \centering
    \includegraphics[width=0.86\textwidth]{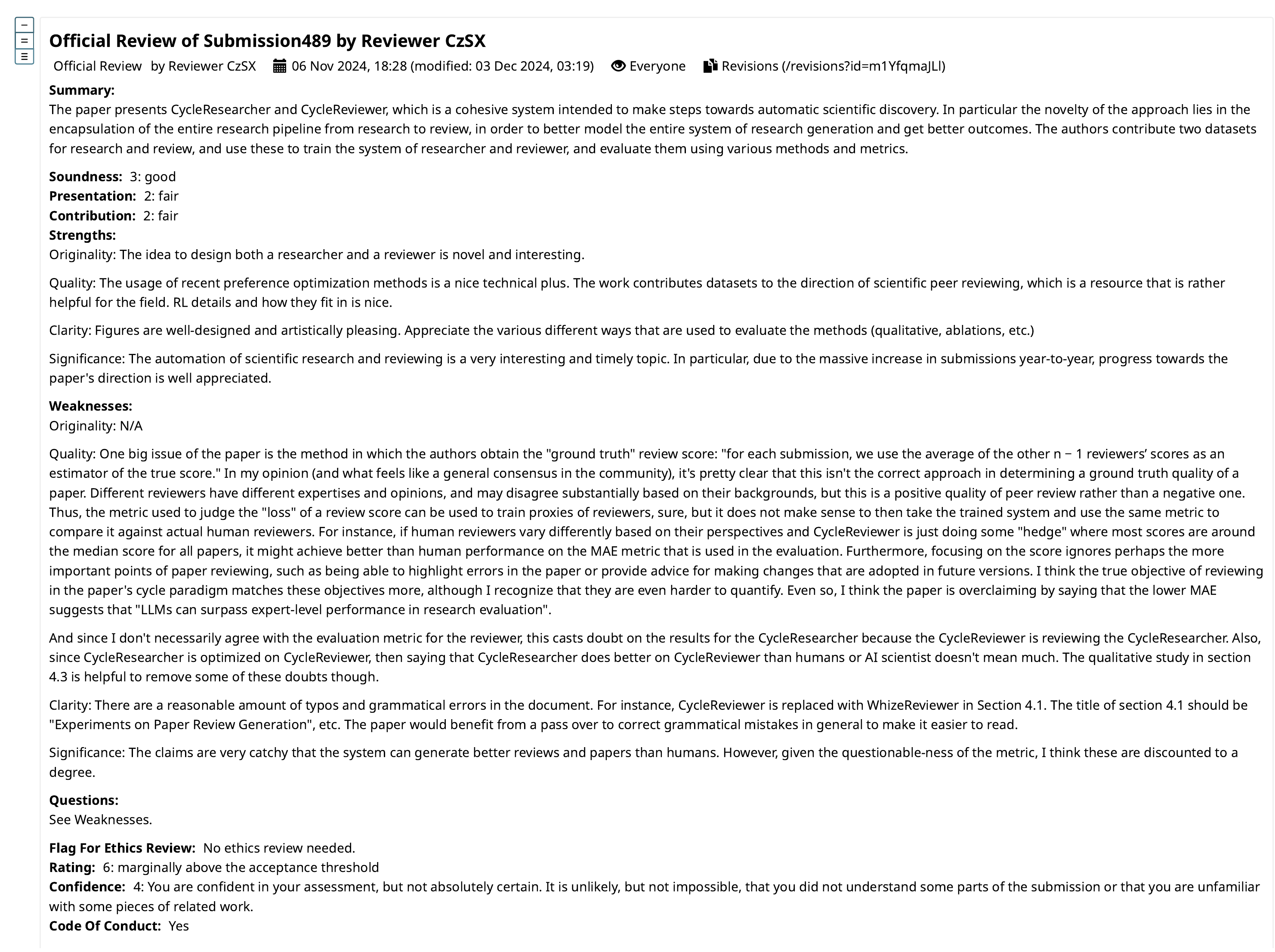}
    \vspace{-0.3cm}
    \caption{The Real-world review comment for CycleResearcher}
    \vspace{-0.3cm}
    \label{fig:r2}
\end{figure*}

\begin{figure*}[t]
    \centering
    \includegraphics[width=0.8\textwidth]{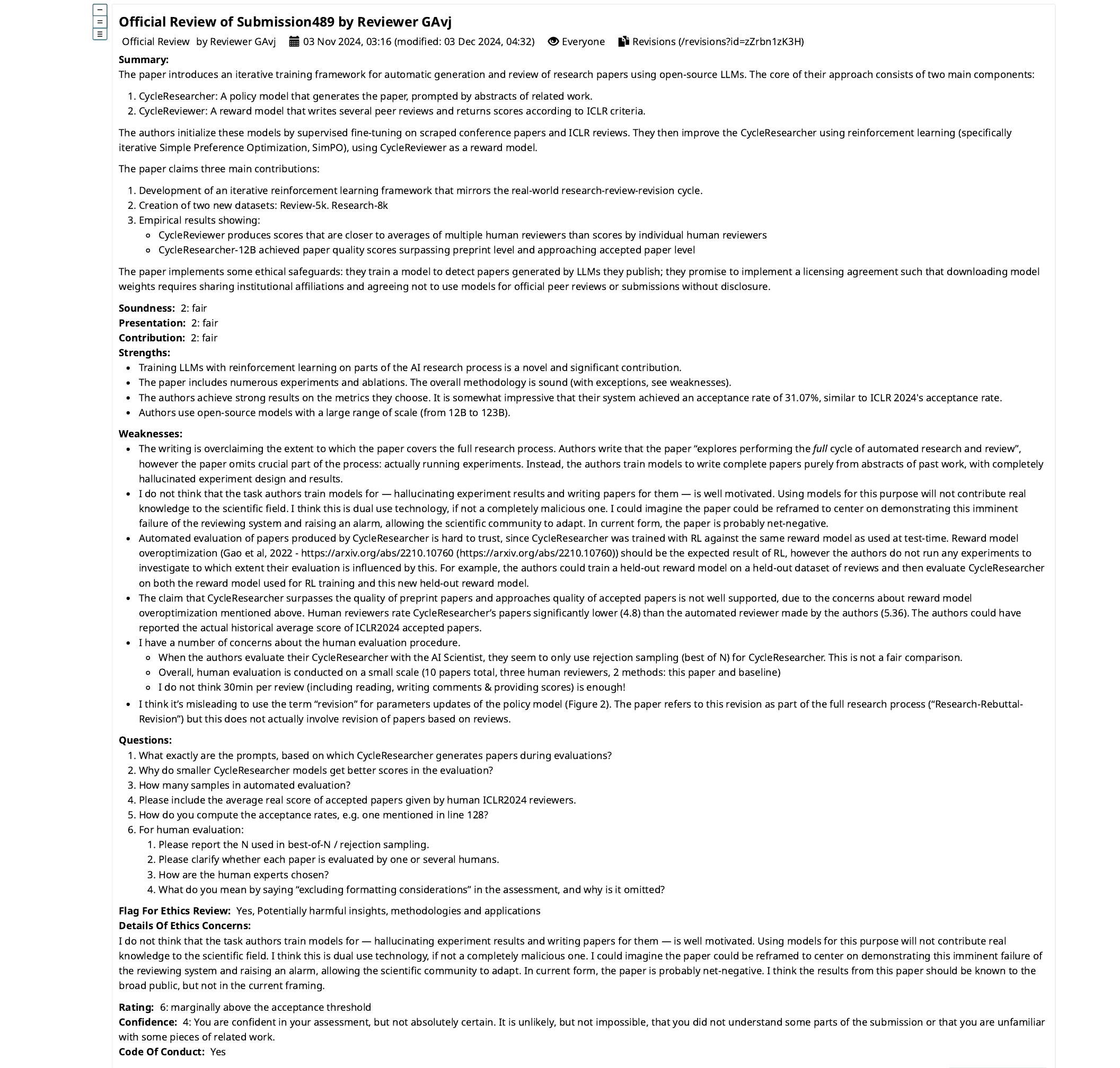}
    \vspace{-0.3cm}
    \caption{The Real-world review comment for CycleResearcher}
    \vspace{-0.3cm}
    \label{fig:r3}
\end{figure*}

\begin{figure*}[t]
    \centering
    \includegraphics[width=0.72\textwidth]{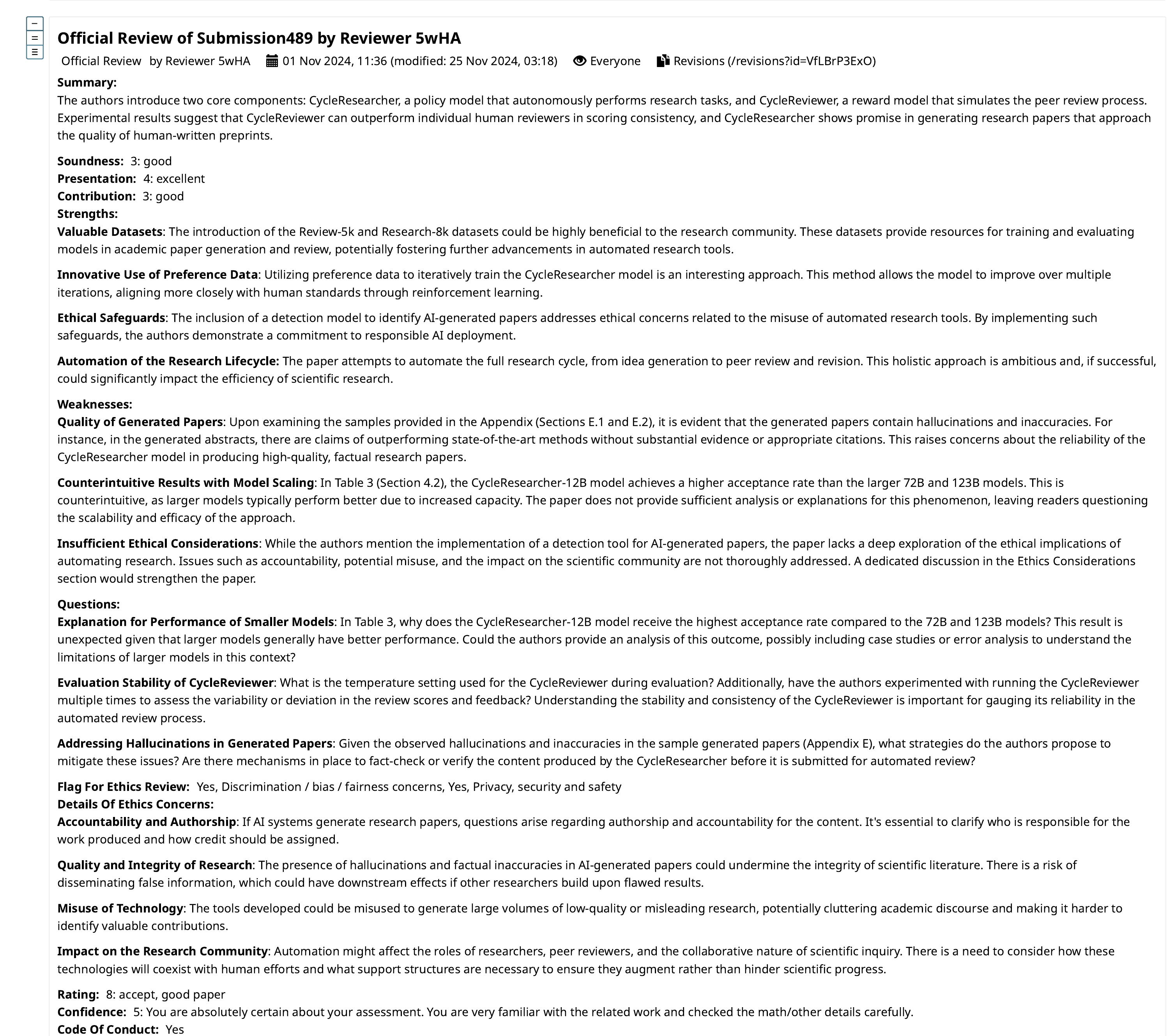}
    \vspace{-0.3cm}
    \caption{The Real-world review comment for CycleResearcher}
    \vspace{-0.3cm}
    \label{fig:r4}
\end{figure*}

To further illustrate the capabilities of DeepReviewer, we present a detailed case study analyzing the Meta-Review generated by DeepReviewer-14B (Best mode) (See in Figure \ref{fig:meta_review}) for the "CycleResearcher" paper\footnote{\url{https://openreview.net/forum?id=bjcsVLoHYs}} \citep{yixuan2024cycleresearcher}, a submission from ICLR 2025 not included in the training dataset. This paper, focusing on automating the research lifecycle with LLMs, received four independent reviews from human experts (Reviewer 7LzG: Figure \ref{fig:r1}, CzSX: Figure \ref{fig:r2}, GAvj: Figure \ref{fig:r3}, and 5wHA: Figure \ref{fig:r4}).  DeepReviewer-14B, operating in its most comprehensive "Best" mode, synthesized these diverse perspectives into a single Meta-Review, aiming to emulate the holistic understanding and critical assessment of a seasoned meta-reviewer.  A preliminary examination reveals a striking alignment between DeepReviewer's Meta-Review and the individual human assessments, both in terms of overall sentiment, identified strengths and weaknesses, and even the final score prediction, which closely mirrors the average human rating. This case study delves deeper into the nuances of this comparison, highlighting both the remarkable capabilities and subtle limitations of DeepReviewer in mimicking expert meta-reviewing.

Comparing the summaries, DeepReviewer accurately captures the core contribution of the "CycleResearcher" paper, emphasizing the novel framework for automating the research lifecycle with LLMs, the two key components (CycleResearcher and CycleReviewer), the iterative reinforcement learning approach (SimPO), and the creation of the Review-5k and Research-8k datasets. This summary resonates strongly with the initial summaries provided by all four human reviewers, each of whom also highlighted these central aspects of the paper.  Furthermore, DeepReviewer's identified strengths mirror the positive aspects recognized by the human reviewers. For instance, the "innovative approach to automating the research lifecycle" echoes Reviewer 7LzG's praise for the "highly innovative" framework and Reviewer 5wHA's acknowledgment of the "Innovative Use of Preference Data" and "Automation of the Research Lifecycle."  The appreciation for the "Review-5k and Research-8k datasets" also aligns with Reviewer 5wHA's explicit mention of "Valuable Datasets" and Reviewer CzSX's comment on the datasets being a "resource that is rather helpful for the field."  Similarly, the recognition of the "CycleResearcher model generates papers with an average quality level close to human-written preprints" echoes Reviewer GAvj's observation that the system "achieved an acceptance rate of 31.07\%, similar to ICLR 2024's acceptance rate" and Reviewer 7LzG’s claim of "papers of quality close to human-written preprints."

The most compelling aspect of DeepReviewer's Meta-Review is its synthesis of weaknesses and corresponding suggestions, demonstrating an ability to identify and consolidate critical concerns raised across different reviewers. DeepReviewer's critique regarding "potential for bias in the training data" and "lack of analysis of diversity" directly addresses concerns implicitly or explicitly raised by reviewers, particularly regarding generalizability and potential limitations of the datasets.  The weakness concerning "computational resources" aligns with Reviewer 7LzG's mention of "Complexity of Implementation" and the need for "significant computational resources."  Similarly, the concern about the "potential for misuse" and the need for "robust safeguards" reflects the ethical considerations raised by Reviewer 5wHA ("Insufficient Ethical Considerations," "Misuse of Technology") and Reviewer GAvj ("Potentially harmful insights, methodologies and applications").  The suggestion for "more details on the specific prompts" and "evaluation criteria" addresses the implicit desire for more clarity on methodology, a common thread in academic reviews.  Finally, the point about "generalizability across different research domains" directly mirrors Reviewer 7LzG's primary "Weakness: Generalizability Across Domains."  This systematic identification and aggregation of weaknesses and suggestions from multiple reviewers showcase DeepReviewer's capacity to perform a nuanced and comprehensive meta-analysis.

While DeepReviewer-14B demonstrates a remarkable ability to synthesize human review insights, it is important to acknowledge potential limitations.  For instance, while DeepReviewer captures the essence of the critiques, the depth of technical understanding in specific areas might not fully match that of a human meta-reviewer deeply versed in the nuances of reinforcement learning or AI ethics.  Furthermore, the Meta-Review, while comprehensive, might lack the subtle nuances and perspectives that a human meta-reviewer could bring to the synthesis process, potentially overlooking more implicit or nuanced concerns expressed in the individual reviews.  However, despite these subtle limitations, DeepReviewer's performance in generating a coherent, insightful, and critically aligned Meta-Review is undeniably impressive.  Crucially, DeepReviewer’s overall rating prediction of 6.0 aligns closely with the average human rating, further validating its ability to not only understand the qualitative aspects of paper evaluation but also to synthesize them into a quantitative judgment consistent with expert consensus. This case study underscores DeepReviewer's potential as a powerful tool for assisting and potentially augmenting the peer review process.

\section{Information About Use Of AI Assistants}

This article has been reviewed by DeepReviewer-14B and revised accordingly based on its review comments.

\end{document}